\documentclass{article} 
\usepackage[final]{colm2026_conference}

\usepackage{microtype}
\usepackage{hyperref}
\usepackage{url}
\usepackage{booktabs}

\usepackage{graphicx}
\usepackage{graphicx} 
\usepackage{makecell}
\usepackage{subcaption}
\usepackage{amsmath, amssymb, amsthm}
\usepackage{mathtools}
\usepackage{multirow}
\usepackage{float}
\usepackage{wrapfig}
\usepackage{adjustbox}
\usepackage{algorithm}
\usepackage{algpseudocode}
\usepackage{booktabs} 

\theoremstyle{plain}

\newtheorem*{corollary*}{Corollary}
\newtheorem*{definition*}{Definition}

\usepackage{iftex}
\usepackage{etoolbox}

\newif\ifarxivhtml
\newcommand{\detecthtmlmode}{
  \ifarxivhtml\else
    \IfFileExists{html.sty}{\arxivhtmltrue}{}
  \fi
}

\detecthtmlmode

\ifarxivhtml
  \newenvironment{smartbox}[1]{\par\vspace{1em}\noindent\textbf{#1.}\quad}{\vspace{1em}\par}
\else
  \usepackage[most]{tcolorbox}
  \newtcolorbox{smartbox}[1]{
    colback=gray!5!white,
    colframe=gray!80!black,
    fonttitle=\bfseries,
    title=#1,
    boxrule=0.5pt,
    arc=2mm,
    left=1mm,
    right=1mm,
    top=1mm,
    bottom=1mm,
    enhanced
  }
\fi

\usepackage{lineno}

\definecolor{darkblue}{rgb}{0, 0, 0.5}
\hypersetup{colorlinks=true, citecolor=darkblue, linkcolor=darkblue, urlcolor=darkblue}

\title{The Generalization Ridge: Information Flow in Natural Language Generation}

\author{
  Ruidi Chang, Chunyuan Deng, Hanjie Chen
  \\
  \\
  Department of Computer Science
  \\
  Rice University
  \\
  Houston, TX 77005, USA
  \\
  \texttt{\{ruidi.chang,chunyuan.deng,hanjie\}@rice.edu}
}

\begin{document}

\ifcolmsubmission
\linenumbers
\fi

\maketitle

\begin{abstract}
Transformer-based language models have achieved state-of-the-art performance in natural language generation (NLG), yet their internal mechanisms for synthesizing task-relevant information remain insufficiently understood. 
While prior studies suggest that intermediate layers often yield more generalizable representations than final layers, how this generalization ability emerges and propagates across layers during training remains unclear.
We propose InfoRidge, an information-theoretic framework, to characterize how predictive information—the mutual information between hidden representations and target outputs—varies across depth during training. Our experiments across various models and datasets reveal a consistent non-monotonic trend: predictive information peaks in intermediate layers—forming a \textbf{generalization ridge}—before declining in final layers, reflecting a transition between generalization and memorization. To further investigate this phenomenon, we conduct a set of complementary analyses that leverage residual scaling and attention patterns to characterize layer-wise functional specialization. We further validate our findings with multiple-token generation experiments, verifying that the observed ridge phenomenon persists across decoding steps. Together, these findings offer new insights into the internal mechanisms of transformers and underscore the critical role of intermediate layers in supporting generalization.\footnote{The code can be found at https://github.com/chili-lab/InfoRidge.}
\end{abstract}

\section{Introduction}

Transformer-based language models generate text by predicting tokens autoregressively, and they have achieved remarkable performance across a wide range of natural-language uses~\citep{vaswani2017attention,dong2022survey}.
Nevertheless, we lack a rigorous understanding of how these models acquire and synthesize task-relevant information during training.

A growing body of research has shown that intermediate layers in deep neural networks often surpass final layers in terms of representational quality and generalization performance~\citep{liu2019linguistic,voita2019bottom,ansuini2019intrinsic,ahrens2023read,uselis2025intermediate}. In language models, intermediate layers often encode richer semantic and more robust features than final layers~\citep{fan2024not,jin2024exploring,skean2025layer}. However, questions still remain: \textbf{\textit{In NLG, how does information evolve across layers during training, and how are different layers of the network functionally organized to support generalization versus memorization?}}

To investigate these questions, we propose \emph{InfoRidge}, an information-theoretic framework, to analyze information flow in language models. Building on matrix-based mutual information estimation~\citep{giraldo2014measures}, our approach quantifies how predictive signals transform across layers during training. 
We center our analysis on two complementary quantities: \textit{predictive information}, defined as the mutual information $I(Z_\ell; Y)$ between the hidden representation $Z_\ell$ at layer $\ell$ and the next-token label $Y$, reflecting how much task-relevant information is preserved; \textit{incremental information gain}, denoted as $I(\Delta Z_\ell; Y)$, where $\Delta Z_\ell$ is the residual change between successive layer $\ell$ and $\ell-1$, measuring the additional predictive information introduced by each transformer block.

Using \emph{InfoRidge}, we uncover a non-monotonic trend: predictive information rises through the early and middle layers, peaks in the upper-middle layers, and then declines in the later layers. We name this phenomenon the \textbf{\textit{generalization ridge}}, where the model encodes the more generalizable task-relevant information. This ridge marks a structural division of labor: intermediate layers concentrate generalizable features that transfer across distributions, while later layers increasingly specialize in task-specific memorization. Incremental information gain further shows that ridge layers introduce the largest increases in predictive information, marking them as key contributors to the emergence of the ridge. This analysis directly connects the information peak to generalizable behavior and clarifies how generalization and memorization are distributed across depth.  

We further extend InfoRidge to multi-token natural language generation. By treating the full output sequence as a structured target and measuring mutual information between hidden states and the long-form target distribution, we show that the generalization ridge persists across decoding steps. Despite increasing mutual information as generation progresses, the layer-wise structure remains stable: predictive information consistently peaks in early-to-intermediate layers and declines in later layers.

To understand the formation of the generalization ridge, we analyze attention patterns and residual scaling coefficients. Attention analysis shows ridge layers attend to tokens that capture broadly useful features, aligning with the information peak. We also investigate the conditions under which the ridge emerges. We also introduce residual scaling coefficients $\beta_\ell$—learnable scalars applied to each residual block with all other weights frozen—as functional probes of layer-wise reliance~\citep{liu2019self, menghani2024laurel}. When evaluated under distribution shift, models have lower reliance on the final layer and shift importance toward the intermediate layers.

\paragraph{Contributions.}
Our work provides a unified perspective on how predictive information is structured across depth in transformer-based language models for natural language generation tasks:

1. Our work tracks the evolution of predictive information throughout training, establishing a clear connection between predictive information flow and generalization. It reveals a non-monotonic peak in the middle layers, which we refer to as the \textbf{\textit{generalization ridge}}. This pattern reflects a meaningful transition in representational focus and aligns with stronger generalization behavior.

2. We introduce InfoRidge, an information-theoretic framework that applies matrix-based mutual information estimation to autoregressive language models to analyze information flow.

3. We analyze attention patterns and residual scaling coefficients, providing causal evidence for the generalization ridge phenomenon.
\section{InfoRidge: information estimation framework}
\label{sec:frame}

\begin{figure*}[t]
    \centering
    \includegraphics[width=\linewidth]{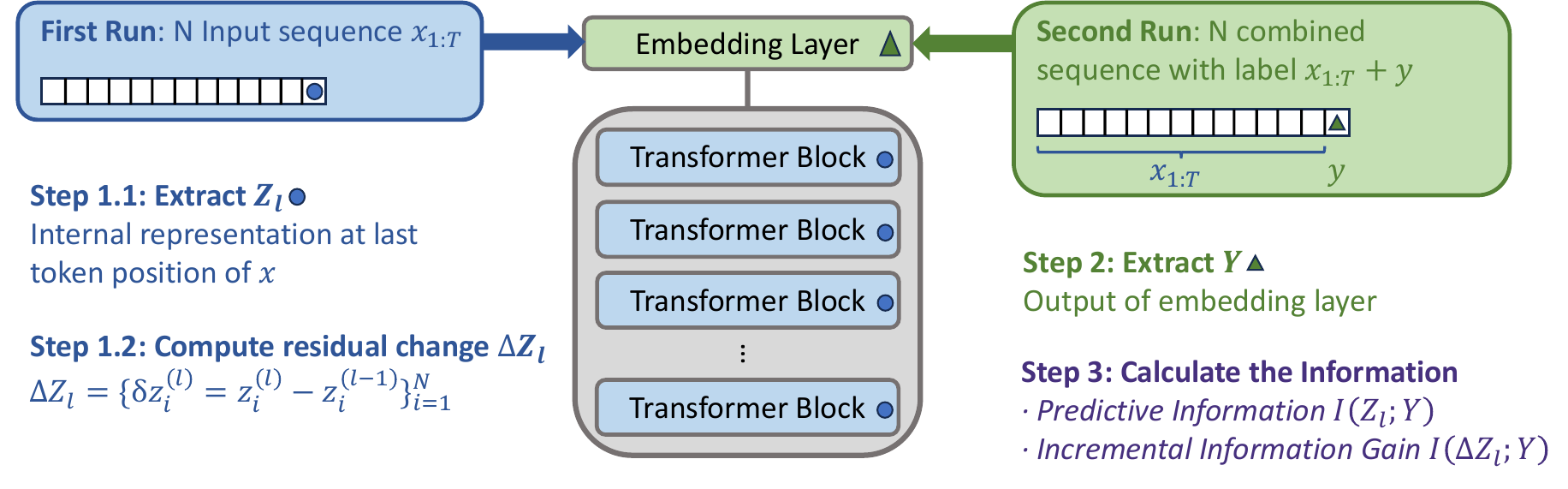}
    \caption{Overview of InfoRidge. (1) Step 1: Extract internal representations at each layer and compute residual changes between successive layers. (2) Step 2: Extract the target token embedding. (3) Step 3: Using these representations, we estimate the layer-wise predictive information $I(Z_{\ell}; Y)$ and the incremental information gain $I(\Delta Z_{\ell}; Y)$, which jointly characterize how information evolves across depth.}
    \label{fig:info-framework}
\end{figure*}

We propose InfoRidge, an information-theoretic framework that uses mutual information to quantify how predictive information propagates through transformer layers during training in NLG.

\paragraph{Motivating Insight.}

Prior work has shown that internal representations in deep neural networks tend to align most closely with the true data distribution at an intermediate layer~\citep{he2024information}. By employing the Wasserstein distance~\citep{villani2008optimal}, this alignment is shown to reach a minimum at a specific depth—referred to as the \emph{generalization funnel layer}. At this point, the \emph{Min Wasserstein Generalization Bound}~\citep{he2024information} ensures that the upper bound on the generalization gap—defined as the expected difference between the population and empirical risks—is minimized. This highlights the critical role of intermediate layers in supporting generalization.

\paragraph{Research Question.}
Despite the insight from prior studies of deep neural networks in classification, it remains unclear whether this generalizes to \emph{Transformer-based language models in natural language generation (NLG)}. This motivates our central question: \textbf{In NLG, how does information evolve across layers during training, and how are different layers of the network functionally organized to support generalization versus memorization?}

\paragraph{Hypothesis.}
Building on the insight, we hypothesize that there exists a specific intermediate transformer layer that encodes the most generalizable representations for next-token prediction, characterized by maximal mutual information with the target label. We refer to this layer as the \textbf{\emph{generalization ridge}}.

\begin{smartbox}{Hypothesis: Generalization Ridge}
There exists an intermediate layer $\ell^* \in \{1, \dots, L\}$ such that the mutual information between the hidden state and the target label peaks at that layer:
\[
\ell^* = \arg\max_{\ell} I(Z_\ell; Y).
\]
We refer to this layer as the \textbf{\emph{generalization ridge}}. This ridge layer aligns most strongly with generalizable features and serves as robust predictors under distribution shift, whereas later layers increasingly specialize in memorization.
\end{smartbox}

\paragraph{Notation and Setup.}

We consider a transformer model with $L$ residual blocks. Given an input sequence $x_{1:T}$ of length $T$, $z^{(\ell)}_i \in \mathbb{R}^d$ denotes the hidden state at the last token position of the $i$th input sequence in layer $\ell$, for $\ell = 1, \dots, L$. For next-token prediction, the ground-truth label is denoted by $y_i \in \mathbb{R}^d$, corresponding to the embedding of the true next token from the vocabulary $\mathcal{V}$. The residual transformation introduced at layer $\ell$ is defined as $\delta z_i^{(\ell)} = z_i^{(\ell)} - z_i^{(\ell-1)}$.

Across a batch of $N$ sequences, we collect the representations:

\[
Z_\ell = \{ z^{(\ell)}_i \}_{i=1}^N\!,\;
Y = \{ y_i \}_{i=1}^N\!,\;
\Delta Z_\ell = \{ \delta z_i^{(\ell)} \}_{i=1}^N,
\]

where all vectors are $\ell_2$-normalized.

\paragraph{InfoRidge Overview.}
To empirically investigate this hypothesis, we propose InfoRidge, an information estimation framework, that characterizes how predictive information evolves across transformer layers. Specifically, we estimate two key quantities:

\begin{itemize}
    \item \textbf{\textit{Predictive Information}} $I(Z_\ell; Y)$: the mutual information between the hidden state at layer $\ell$ and the target token. This quantity measures how much information about the true next token is contained in the layer’s full representation. A high value indicates that the layer encodes a strong and direct signal relevant to the prediction task.
    \item \textbf{\textit{Incremental Information Gain}} $I(\Delta Z_\ell; Y)$: the information introduced by the residual transformation at layer $\ell$, where $\Delta Z_\ell = Z_\ell - Z_{\ell-1}$. This captures the additional predictive signal gained through the residual transformation at layer $\ell$, isolating how much new task-relevant information is introduced on top of the previous layer’s representation.
\end{itemize}

Together, these metrics allow us to track both the accumulation and transformation of task-relevant information throughout the network.

\paragraph{Computational Flow Overview.}
Figure~\ref{fig:info-framework} illustrates the workflow used to extract intermediate representations for information analysis. In the first forward pass, a batch of $N$ input sequences $x_{1:T}$ is fed into the transformer to obtain hidden states at each layer. From these, we extract the final-token representations $Z_\ell$ and compute the residual changes $\Delta Z_\ell$ by differencing consecutive layer outputs. In the second forward pass, each input is concatenated with its ground-truth next token $y$, and we extract the corresponding label embedding $Y$ from the output of the embedding layer. These representations are then used to compute two information-theoretic quantities: the \textit{Predictive Information} $I(Z_\ell; Y)$, and the \textit{Incremental Information Gain} $I(\Delta Z_\ell; Y)$. We estimate both $I(Z_\ell; Y)$ and $I(\Delta Z_\ell; Y)$ using Equations~\ref{eq:entropy} and~\ref{eq:mutual}, detailed below.

\paragraph{Matrix-Based Mutual Information.}
We apply the matrix-based framework~\citep{giraldo2014measures} to estimate mutual information. Let $\mathcal{U}$ be a random variable, from which we draw a set of vectors $U = \{\mathbf{u}_i\}_{i=1}^N \subset \mathbb{R}^d$. A positive-definite Gram matrix $G_U \in \mathbb{R}^{N \times N}$ is computed using a Gaussian kernel $\kappa$ with bandwidth set to 1 and the matrix is then trace-normalized to satisfy $\operatorname{tr}(G_U) = 1$. To assess robustness with respect to kernel selection, we additionally test the Laplacian and Polynomial kernels. While these kernels introduce quantitative shifts in the magnitude, the trend remains unchanged, see Appendix~\ref{appendix:fullresults}. The matrix-based Rényi entropy (with order $\alpha = 1$) is then given by:
\begin{equation}
    H(\mathcal{U}) \approx H(U) = -\operatorname{tr}(G_U \log G_U).
    \label{eq:entropy}
\end{equation}
Specifically, $G_U$ is constructed with entries
\(
(G_{U})_{ij} = \exp\left( -\frac{\| u_i - u_j \|^2}{2\sigma^2} \right)
\) and then trace-normalized.

Let $G_{\mathcal{U}}$ and $G_{\mathcal{V}}$ be the trace-normalized Gram matrices for two random variables $\mathcal{U}$ and $\mathcal{V}$, respectively. The mutual information between them is computed as:
\begin{equation}
\begin{aligned}
I(\mathcal{U}; \mathcal{V}) 
&\approx I(G_{\mathcal{U}}; G_{\mathcal{V}}) \\
&= H(G_{\mathcal{U}}) + H(G_{\mathcal{V}}) - H(G_{\mathcal{U}} \circ G_{\mathcal{V}}),
\label{eq:mutual}
\end{aligned}
\end{equation}
where ``$\circ$'' denotes the Hadamard product. Mathematical details are in Appendix~\ref{appendix:math}.
\section{Experimental setup}
\label{sec:experiments}

\paragraph{Models.}
We evaluate four models: \textsc{GPT--2 Small} (117M)~\citep{radford2019language}, \textsc{GPT--2 Medium} (345M)~\citep{radford2019language}, \textsc{Qwen--2.5 0.5B}~\citep{qwen2}, and \textsc{LLaMA~3.1 8B}~\citep{meta_llama_3.1_8b}. For the experiments in Section~\ref{sec:results}, all models are fine-tuned on their respective downstream tasks.

\paragraph{Datasets.}
We assess model behavior across four tasks casted into NLG problems: CLUTRR~\citep{sinha2019clutrr}, a relational reasoning benchmark; ECQA~\citep{aggarwal2021explanations}, a commonsense QA benchmark; CNN/DailyMail~\citep{see-etal-2017-get,NIPS2015_afdec700}, a summarization benchmark and Synthetic Arithmetic, a controlled dataset designed to disentangle task-relevant signal from noise. Dataset and implementation details are in Appendix~\ref{appendix:dataset} and~\ref{appendix:prompt}.

\textit{Synthetic Arithmetic Dataset Construction.} We construct a synthetic dataset to separate signal learning from noise memorization. Each example is a sequence of 10 elements, where the signal follows an arithmetic progression modulo $K$, computed as $s_t = (s_0 + t \cdot d) \bmod K$ with $s_0 \in [0, K{-}1]$ and $d \in [1, K{-}1]$. Each element takes the form \texttt{S\{signal\}\_N\{noise\}}, where noise is sampled from $\mathcal{U}_\text{int}(0, \texttt{noise\_range-1})$ ($\mathcal{U}_\text{int}$ denotes the uniform distribution). The model is trained to predict the signal value of the final (10th) element using the preceding elements as input context. For example, with $K = 5$, $s_0 = 1$, and $d = 2$, a sample input might be \texttt{S1\_N42 S3\_N77...S2\_N37}, with the target signal being \texttt{4}. By varying $K$, we can induce structured distribution shifts.

\section{Generalization Ridge: layer-wise mutual information training dynamics}
\label{sec:results}

Understanding how layers encode task-relevant information is key to uncovering the internal mechanisms that support generalization in language models. In this section, we trace the evolution of two complementary forms of mutual information—\textit{Predictive Information} and \textit{Incremental Information Gain}—across transformer depth and training time, revealing a structure in information flow and highlighting the generalization–memorization trade-off.

\subsection{Predictive Information: information peaks at intermediate layers}

We investigate how \textit{predictive information}—defined as the mutual information between hidden representations and target labels—evolves across the depth of transformer models. Specifically, for each layer $\ell$, we compute the matrix-based mutual information $I(Z_\ell; Y)$ between the hidden state $Z_\ell$ and the next-token ground truth $Y$. This quantity measures how much task-relevant signal is retained in the representation as it propagates through the network. By tracing $I(Z_\ell; Y)$ across layers, we obtain a layer-wise trajectory of information flow, which reveals not only where predictive content is preserved but also how it is transformed or diminished as the model processes.

\begin{figure}[h]
    \centering
    \begin{subfigure}[b]{0.48\linewidth}
        \centering
        \includegraphics[width=\linewidth]{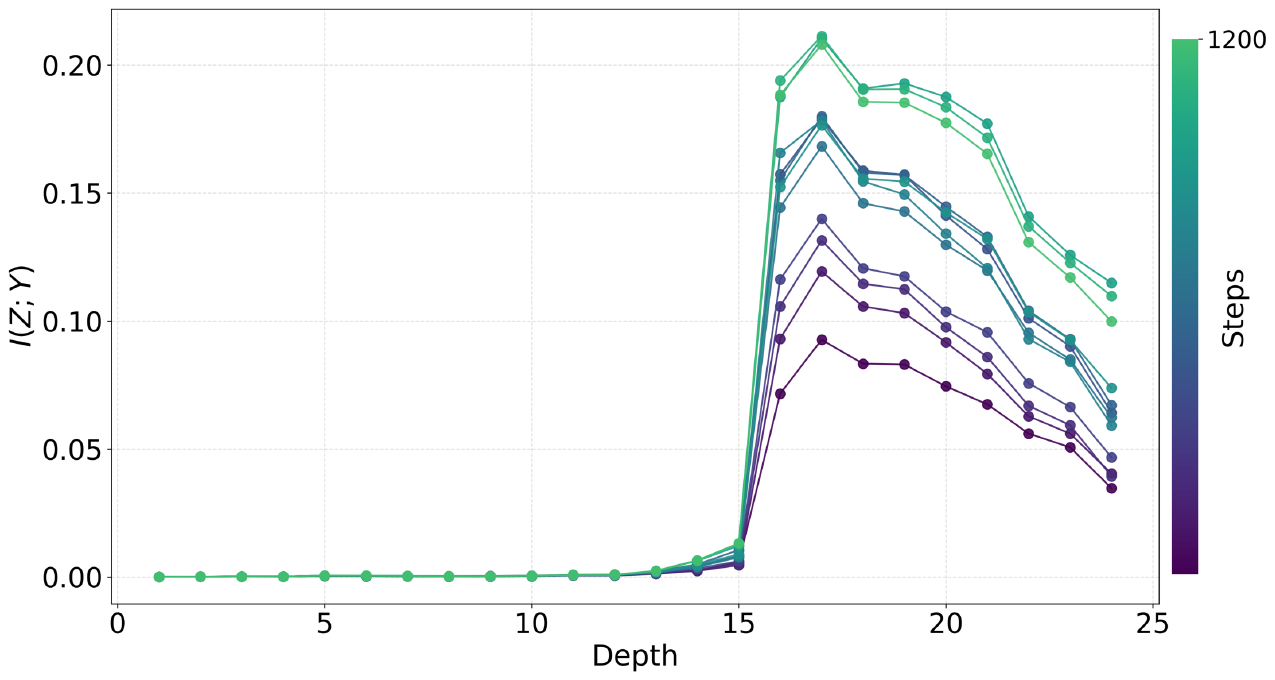}
        \caption{Qwen-2.5-0.5B on \textsc{ECQA}}
        \label{fig:qwen}
    \end{subfigure}
    \hfill
    \begin{subfigure}[b]{0.48\linewidth}
        \centering
        \includegraphics[width=\linewidth]{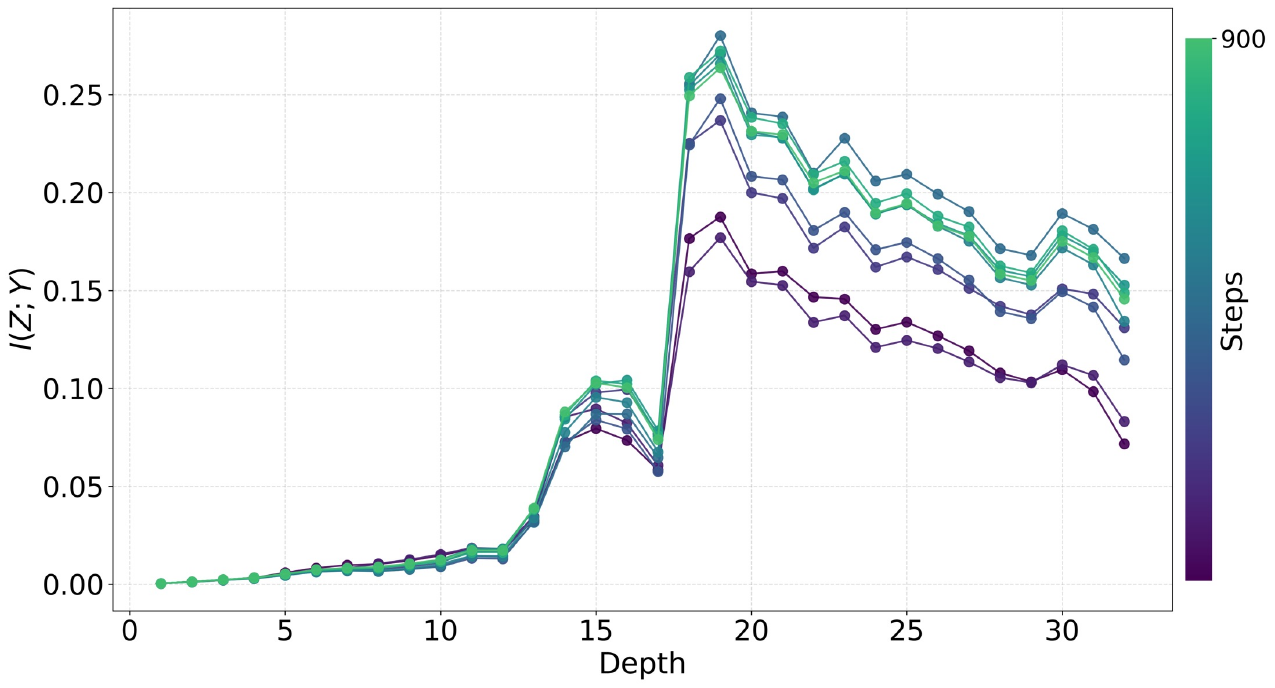}
        \caption{LLaMA-3.1-8B on \textsc{ECQA}}
        \label{fig:llama}
    \end{subfigure}
    \caption{Evolution of predictive information \(I(Z_\ell;Y)\), with lighter curves indicating later training steps. Each curve exhibits a three-phase trend: early layers rise, mid layers peak, and late layers decline.}
    \label{fig:stepwise-mi-all}
\end{figure}

Figure \ref{fig:stepwise-mi-all} tracks the trajectory of the predictive information $I(Z_\ell;Y)$ between the hidden representation at depth $\ell$ and the target label $Y$ throughout training, while Table \ref{tab:layer_eval_summary} reports the downstream accuracy obtained when we early exit after a given layer.\footnote{The in distribution split is generated with $K_{\text{id}}{=}13$; out of distribution splits use a uniformly–sampled $K_{\text{ood}}\in\{5,\dots,25\}\setminus\{13\}$.} Additional results are in Appendix~\ref{appendix:fullresults}.

\paragraph{Three-phase Information Dynamics.}
Across models and tasks, predictive information curves exhibit a consistent \emph{three-phase} pattern:

\textit{Progressive Accrual (early layers).} In the initial layers, $I(Z_\ell;Y)$ gradually increases, corresponding to basic feature extraction without substantial task-level comprehension. This aligns with the near-zero accuracy observed in Table~\ref{tab:layer_eval_summary} for these layers.

\textit{Information Peak (intermediate layers).} $I(Z_\ell;Y)$ continues to rise through the mid-to-upper layer, typically peaking before the final few blocks. For the GPT-2 Small on Synthetic Arithmetic dataset, the peak reaches $I\!\approx\!0.27$ in the 10th layer, coinciding with a jump from $\approx 0\%$ to $57\%$ OOD accuracy. These layers appear to play a critical role in synthesizing abstract features that are essential for generalization.

\begin{wraptable}[20]{r}{0.5\textwidth}
\begin{center}
\adjustbox{max width=0.5\textwidth}{
\begin{tabular}{lcccc}
\toprule
\textbf{Layer} & \(\mathbf{I(Z;Y)}\) & \multicolumn{3}{c}{\textbf{Test Accuracy (\%)}} \\
\cmidrule(lr){3-5}
& & \textbf{All} & \textbf{In-Dist.} & \textbf{Out-Dist.} \\
\midrule
Layer 1  & 0.0002 & 0.00 & 0.00 & 0.00 \\
Layer 2  & 0.0003 & 0.00 & 0.00 & 0.00 \\
Layer 3  & 0.0005 & 0.00 & 0.00 & 0.00 \\
Layer 4  & 0.0013 & 0.00 & 0.00 & 0.00 \\
Layer 5  & 0.0043 & 0.00 & 0.00 & 0.00 \\
Layer 6  & 0.0185 & 0.00 & 0.00 & 0.00 \\
Layer 7  & 0.0416 & 2.67 & 0.20 & 5.13 \\
Layer 8  & \textbf{0.1883} & 17.78 & 0.20 & \textbf{35.37} \\
Layer 9  & 0.2301 & 20.90 & 1.47 & 40.33 \\
Layer 10 & \textbf{0.2721} & 55.15 & 53.40 & \textbf{56.90} \\
Layer 11 & 0.2475 & 65.92 & 79.17 & 52.67 \\
Layer 12 & \textbf{0.0188} & 71.58 & \textbf{93.17} & 50.00 \\
\bottomrule
\end{tabular}}
\end{center}
\caption{Information dynamics and layer-wise performance (GPT-2-S, Syn). OOD performance declines beyond the ridge.}
\label{tab:layer_eval_summary}
\end{wraptable}

\textit{Representational Compression (final layers).} Beyond the peak, $I(Z_\ell;Y)$ decreases, even as in-distribution accuracy approaches $93\%$. The simultaneous drop in OOD accuracy indicates that the final layers tend to memorize training patterns, sacrificing generalization ability.

\paragraph{Generalization Ridge: Generalization-Memorization Trade-off.}
The pronounced ``information funnel'' around intermediate layers reflects a key trade-off between generalization and memorization, which we term the ``\textbf{generalization ridge}''. These layers maximize task-relevant information for generalization, while deeper layers increasingly compress and specialize representations, enhancing in-distribution memorization but reducing robustness. This positions intermediate layers as critical control points for managing this trade-off.

\paragraph{Semantically Important Attention Peaks Where Information Peaks.}

This discussion aims to provide an intuitive illustration of our generalization ridge hypothesis, highlighting how attention to task-relevant signal tokens shifts across layers (the generalizable information). Specifically, we computed average signal attention across layers, identified the layer with peak signal attention, and compared it to final-layer signal and last-token attention, results are in Table~\ref{tab:atte-signal}. Signal tokens are defined as task-relevant tokens that the model must focus on to solve the task—for the synthetic dataset, these are tokens that appear after the character ‘S’; for CLUTRR, kinship-related entities; and for ECQA, the token corresponding to the correct answer option. This allows us to quantify where in the network attention to semantically important information is concentrated. Our findings show a pattern that supports our  hypothesis: (1) Mid-to-late layers peak in signal attention, coinciding with the predictive information ridge (Figure~\ref{fig:stepwise-mi-all}), indicating where generalizable representations are strongest. (2) Final layers show reduced attention to signal tokens, suggesting a shift toward memorization to specific data points rather than predictive information abstraction. For example, in Qwen-2.5-0.5B on ECQA, signal attention peaks at Layer 17 (0.2307) but drops to 0.0104 in the final layer, where last-token attention dominates (0.2169).

\begin{table}[h]
\begin{center}
\resizebox{\columnwidth}{!}{
\begin{tabular}{l l c c c c c}
\toprule
\textbf{Model} & \textbf{Dataset} & 
\textbf{Avg. Attn (All)} & 
\textbf{Avg. Attn (Signal)} & 
\textbf{Layer w/ Highest Avg. Signal} & 
\textbf{Final Avg. Signal} & 
\textbf{Final Avg. Last} \\
\midrule
Qwen-2.5-0.5B & ECQA      & 0.0229 & 0.0317 & 17 (0.2307) & 0.0104 & 0.2162 \\
LLaMA-3.1-8B  & ECQA      & 0.0224 & 0.0394 & 21 (0.1595) & 0.0094 & 0.5073 \\
GPT-2 Small   & Synthetic & 0.0222 & 0.0409 & 8 (0.0739)  & 0.0555 & 0.0378 \\
GPT-2 Medium  & CLUTRR    & 0.0080 & 0.0078 & 18 (0.0243) & 0.0112 & 0.1576 \\
\bottomrule
\end{tabular}
}
\end{center}
\parbox{\linewidth}{\raggedright \footnotesize *We remove the first token attention score to mitigate attention sink effects.}
\caption{Average attention statistics: (1) average attention scores over all tokens, (2) average attention to signal tokens, (3) the maximum signal attention and its corresponding layer, (4) signal token attention in the final layer and (5) last token attention in the final layer.}
\label{tab:atte-signal}
\end{table}

\subsection{Incremental Information Gain: information concentrates at intermediate layers}

\begin{wrapfigure}{r}{0.5\textwidth}
    \centering
    \includegraphics[width=\linewidth]{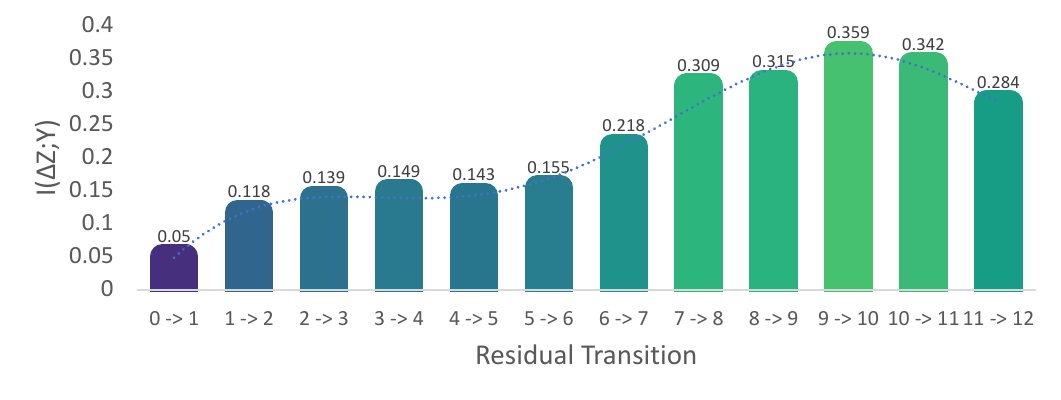}
    \caption{Middle blocks are key to encoding generalizable information (GPT-2-S, Synthetic).}
    \label{fig:delta_z}
\end{wrapfigure}

To understand how information accumulates across the network, we compute \textit{Incremental Information Gain} ($I(\Delta Z_\ell; Y)$)—the mutual information between each residual transition and the target label embedding. As shown in Figure~\ref{fig:delta_z}, the resulting layer-wise gains reveal that intermediate layers yield the highest information increases. This concentration of information gain further underscores their central role in encoding those task-relevant features that are essential for supporting generalization. The Incremental Information Gain analysis reveals a clear pattern: 
middle transformer blocks are key to encoding generalizable task-relevant information, thereby forming a \textit{generalization ridge}. 
Conversely, later layers contribute little additional predictive signal, and in some cases, actively reduce alignment with the target embeddings. This diminishing contribution in later layers may suggest a shift from general reasoning to memorization of training-specific patterns. This trend underscores a fundamental trade-off in transformer training dynamics. Additional results are in Appendix~\ref{appendix:fullresults}.

\paragraph{Residual Scaling Dynamics via Learnable \texorpdfstring{$\beta$}{β} Coefficients.}

\begin{wrapfigure}[15]{r}{0.5\textwidth}
    \centering
    \includegraphics[width=\linewidth]{figures/llama_ecqa_betafig.pdf}
    \caption{Residual scaling coefficients $\beta_\ell$ across all layers. Curve shows the mean across 5 independent runs, and the shaded region denotes 1-$\sigma$ error bar. LLaMA on ECQA: training on OOD data improves OOD performance from $0.0\%$ to $21.1\%$.}
    \label{fig:beta}
\end{wrapfigure}

To deepen our understanding of the Generalization Ridge hypothesis, we examine how modulating the contribution of individual transformer blocks affects information flow.

We introduce a \textit{residual scaling mechanism} with learnable scalar coefficient parameters, inspired by prior work on adaptive residual modulation~\citep{liu2019self, menghani2024laurel}. Transformer architectures inherently employ residual connections to iteratively refine representations. To isolate and quantify the contribution of each transformer block, we scale these residual connections with layer-specific scaling factors $\beta_{\ell} \in \mathbb{R}_{\ge 0}$, with definition below. Each $\beta_\ell$ controls the strength of the residual contribution from layer $\ell$, enabling the model to adaptively emphasize or suppress specific blocks:
\[
z^{(\ell)} = z^{(\ell - 1)} + \beta_{\ell} \cdot \text{block}^{(\ell)}(z^{(\ell - 1)}), \qquad \beta_{\ell} \in \mathbb{R}.
\]

We freeze model weights and optimize only the residual scaling coefficient parameters $\beta_{\ell}$, which are initialized to 1. These scalars are trained separately on the (a) in-distribution (ID) split and (b) out-of-distribution (OOD) split. Since no other parameters are updated, the learned $\beta_{\ell}$ serve as a direct diagnostic of the extent to which each layer should be amplified or attenuated to suit the data regime, revealing which layers remain stable across regimes and which adapt strongly to distributional shifts.

As shown in Figure~\ref{fig:beta}, optimizing the residual scaling coefficient parameters on in-distribution data assigns relatively higher residual weights to the final transformer layers, indicating that in-distribution performance benefits from increased reliance on late-layer residual contributions. This observation suggests that these deeper layers capture refinements that are closely tied to the training distribution. In contrast, under out-of-distribution (OOD) settings, the learned coefficients systematically have relative lower contributions of the final layers compared to the ID settings, revealing that improved generalization is associated with downweighting late-stage residual signals. When train on OOD, intermediate layers are upweighted and OOD performance improves. Additional results are in Appendix~\ref{appendix:fullresults}. This pattern supports the interpretation that late-layer residuals become increasingly specialized in memorized patterns tied to the training distribution. Together, these findings provide empirical support for the generalization ridge hypothesis, demonstrating that information flow in transformers reflects a trade-off between generalizable signals and memorized, distribution-specific features.

\section{Extension to multi-token scenario}

We further evaluate information flow under multi-token outputs using the CNN/DailyMail summarization dataset (1.0.0, test) \citep{see-etal-2017-get,NIPS2015_afdec700}. We evaluate two pretrained models: Qwen-2.5-0.5B and LLaMA-3.1-8B, generating 50 tokens per sample.

\begin{wrapfigure}[17]{r}{0.5\textwidth}
    \begin{center}
    \includegraphics[width=\linewidth]{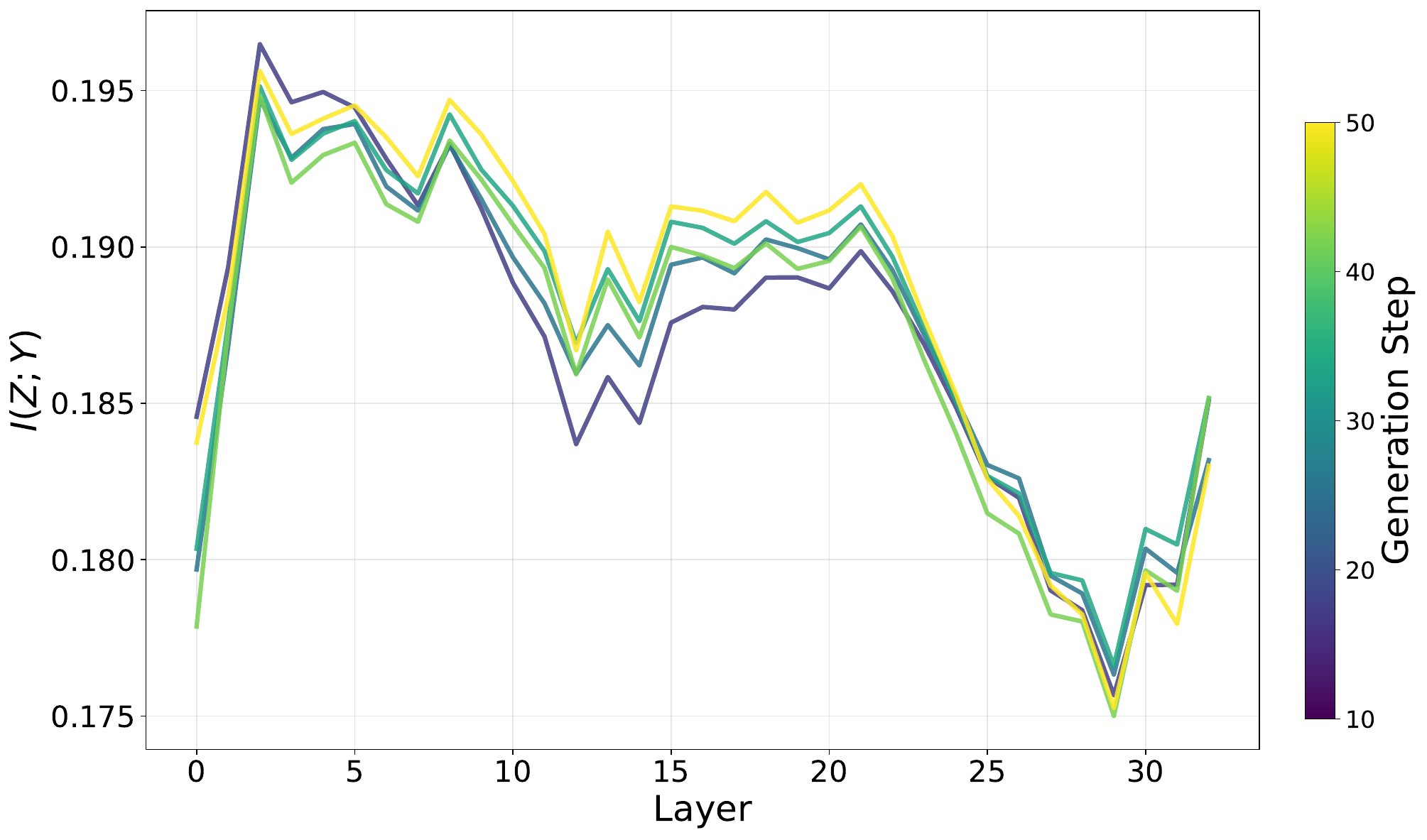}
    \caption{$I(Z_\ell^{(t)} ; Y)$ across layers (x-axis) and generation steps (right color bar), the y-axis shows mutual information values. Values are plotted every ten steps.}
    \label{fig:multi_main}
    \end{center}
\end{wrapfigure}

\paragraph{Kernel Extension for Variable-Length Sequences.}
To handle multi-token targets, we extend our matrix-based mutual information (MI)
estimation to variable-length sequences.
Let $Y_i = \{ y_i^{(t)} \}_{t=1}^{T_i}$ denote the target sequence for sample $i$.
We employ the following kernel to compute sample-to-sample similarity:
\[
K(Y_i, Y_j)
= \frac{1}{T_i T_j}
\sum_{t=1}^{T_i}
\sum_{t'=1}^{T_j}
\kappa\!\left( y_i^{(t)}, y_j^{(t')} \right),
\]
where $\kappa(\cdot, \cdot)$ is a Gaussian kernel.
This formulation aggregates all pairwise token similarities between two target sequences into a single scalar, capturing the overall alignment between two token distributions.
We quantify how much information each layer carries about the gold answer by computing $I(Z_\ell^{(t)} ; Y)$ in Figure~\ref{fig:multi_main}, where $Z_\ell$ denotes the hidden state at the last token position at decoding step $t$ and $Y$ represents the full target sequence. 

\paragraph{Mutual information across generation steps.}
We observe several consistent patterns in the mutual information profiles across generation steps.
First, the overall magnitude of $I(Z_\ell^{(t)}; Y)$ increases as generation progresses: later decoding steps tend to exhibit higher mutual information than earlier ones.
This reflects the accumulation of information in the hidden representations as the model conditions on an increasingly informative prefix composed of its previously generated tokens.
As generation progresses, the hidden state progressively approach the target sequence distribution.
Second, despite differences in magnitude, the layer-wise structure of the mutual information curves remains consistent across generation steps.
In particular, $I(Z_\ell^{(t)}; Y)$ peaks at early-to-intermediate layers and gradually declines toward later layers.
The concentration of mutual information in early-to-intermediate layers reflects their role in forming task-relevant representations under autoregressive decoding.
These layers integrate information from the input and the evolving context into relatively abstract features that are shared across decoding steps, such as semantic content and task-level structure.
While these representations are critical for producing the next output token, later layers increasingly compress information into a lower-entropy, leading to reduced mutual information with the target sequence.
\section{Related work}
\label{sec:relate}
Understanding how information is encoded and transformed across layers has been studied through probing classifiers~\citep{alain2016understanding}, attention flow~\citep{vig2019analyzing}, and information-theoretic approaches such as the information bottleneck~\citep{shwartz2017opening}, mutual information estimation~\citep{goldfeld2019estimating}, and matrix-based entropy~\citep{giraldo2014measures}, offering different lenses to quantify representational capacity, abstraction, and invariance across layers.

A growing body of research has shown that intermediate layers in deep networks often outperform final layers in terms of representational quality and task performance~\citep{ansuini2019intrinsic,yosinski2014transferable,uselis2025intermediate,ahrens2023read,ando2023use}. In language models, mid-depth layers tend to capture richer semantic or robust features than output layers~\citep{liu2019linguistic,voita2019bottom,jin2024exploring,fan2024not}. Transformer representations have long been observed to follow a structured progression from syntactic to semantic information, as shown by classical probing studies on linguistic knowledge and the reconstruction of the NLP pipeline~\citep{liu2019linguistic,tenney2019bert}. These findings challenge the assumption that deeper layers always yield better representations. More recent work further strengthens this observation by explicitly characterizing the role of middle-to-upper layers. 
\citet{queipo2026attention} connect the emergence of attention sinks to representational compression in intermediate layers and propose a ``Mix--Compress--Refine'' view of information flow across model depth.
The semantic hub hypothesis suggests that these layers act as convergence points for abstract semantic representations, while other studies identify an abstraction phase in deeper layers~\citep{wu2024semantic,cheng2024emergence}. 
Earlier probing analyses similarly reveal a structured progression of representations across layers~\citep{tenney2019bert,sajjad2022analyzing}, and layer pruning studies further highlight their non-uniform functional roles~\citep{sajjad2023effect}. Earlier probing analyses similarly reveal a structured progression of representations across layers~\citep{tenney2019bert,sajjad2022analyzing}, and layer pruning studies further highlight their non-uniform functional roles~\citep{sajjad2023effect}.
Complementing these analyses, \citet{ng2026rys} shows that selectively repeating a contiguous block of intermediate transformer layers at inference time can improve model performance without modifying the model weights, providing additional intervention-based evidence that layers differ substantially in their functional contributions.

This pattern holds across settings such as transfer learning~\citep{yosinski2014transferable}, continual learning~\citep{ahrens2023read}, and out-of-distribution generalization~\citep{uselis2025intermediate}. Furthermore, recent work has evaluated representation quality using entropy, curvature, and invariance~\citep{skean2025layer}, while other studies have examined embedding drift and representational geometry~\citep{merchant2020happens,dar2022analyzing}, analyzed memorization and factual recall~\citep{haviv2022understanding,yu2023characterizing}, and introduced causal perspectives on layer importance through mediation analysis and targeted interventions~\citep{vig2020causal,meng2022locating}. Training-dynamics studies investigated how earlier models develop and refine semantic features across depth, providing an additional perspective on the evolution of layer-wise representations~\citep{merchant2020happens,kumar2023grokking}. Correlational probes (e.g., linear probes~\citep{alain2016understanding}) measure only whether a feature can be decoded from a representation, which reflects correlation but not causal influence. In contrast, causal methods intervene on internal activations to test how changing a component alters the model’s prediction, thereby identifying true causal contribution rather than mere feature presence. Our residual-scaling approach aligns with this causal perspective at layer level.

However, the underlying causes and functional role of this phenomenon remain only partially understood, motivating further investigation. Our work addresses this gap by tracing the evolution of predictive information throughout training and establishing a clear connection between predictive information flow and generalization. We reveal a consistent non-monotonic peak in the middle layers—termed the \emph{generalization ridge}—that reflects a meaningful transition in representational focus and aligns with stronger generalization behavior. Additionally, unlike prior work focused on classification tasks, we extend the analysis of information flow to generation tasks, enabling us to understand how generalization and memorization dynamics evolve for natural language generation setting from an information-theoretic perspective.
\section{Conclusion}
\label{sec:conclu}

We introduce InfoRidge, an information-theoretic framework designed to trace and quantify how information evolves across layers in transformer-based language models for natural language generation. By estimating both predictive information and incremental information gain, we systematically characterize the layerwise dynamics of information flow, offering a principled view of how signals are refined, amplified, or diminished as they propagate through the network. Our findings reveal a consistent \textit{generalization ridge} emerging in intermediate layers, where mutual information between the hidden representation and the target label reaches its peak before gradually declining. This phenomenon reflects a fundamental trade-off between generalization and memorization as information flows deeper into the model. A similar pattern is observed in the multi-token setting. Attention analysis and residual scaling experiments offer complementary causal evidence supporting this observation. Taken together, these findings position InfoRidge as a comprehensive framework for diagnosing how language models internally manage information during natural language generation, while also revealing the structural mechanisms that govern the balance between generalization and memorization.
\section*{Acknowledgments}
This project was supported by the Rice Ken Kennedy Institute Research Award No. 1081025 and the U.S. National Institutes of Health under Award Number OT2OD038051.
The content is solely the responsibility of the authors and does not necessarily represent the official views of the National Institutes of Health.

\bibliography{colm2026_conference}
\bibliographystyle{colm2026_conference}

\appendix
\clearpage


\section{Mathematical details for matrix-based information estimation and theoretical foundations}
\label{appendix:math}

We employ the matrix-based Rényi entropy ~\citep{giraldo2014measures} to estimate mutual information between representations and labels, leveraging kernel Gram matrices to capture sample similarity.

\subsection{Matrix-based entropy estimation}

Let $U = \{u_i\}_{i=1}^N \subset \mathbb{R}^d$ denote $\ell_2$-normalized representations obtained from a specific transformer layer, a residual update, or the embedding of the target label. A Gaussian kernel Gram matrix $G_U \in \mathbb{R}^{N \times N}$ is constructed as:
\(
(G_U)_{ij} = \exp\left( -\frac{\| u_i - u_j \|^2}{2 \sigma^2} \right),
\)
with bandwidth $\sigma = 1$. The matrix is then trace-normalized to ensure $\text{tr}(G_U) = 1$.

The matrix-based Rényi entropy of order $\alpha = 1$ is defined as:

$$
H(U) = - \text{tr}(G_U \log G_U).
$$

This expression can be interpreted in terms of the eigenvalue spectrum $\{ \lambda_k \}$ of $G_U$, since $G_U$ is positive semi-definite and trace-normalized:

$$
H(U) = - \sum_{k=1}^{N} \lambda_k \log \lambda_k.
$$

The entropy thus reflects the dispersion of the eigenvalues. A more uniform spectrum (i.e., higher entropy) suggests more diversity in the representation space, while a sharply peaked spectrum (i.e., low entropy) indicates redundancy or compression.

\textit{Polynomial Kernel.}
Given a set of normalized representations 
$U = \{u_i\}_{i=1}^N$, 
the polynomial kernel Gram matrix is defined as
\(
    (G^{\mathrm{poly}}_U)_{ij}
    = \left( u_i^\top u_j + c_0 \right)^{p},
\)
where $p$ is the polynomial degree and $c_0$ is a constant bias term.
\textit{Laplacian Kernel.}
Given the same representation set $U = \{u_i\}_{i=1}^N$, 
the Laplacian kernel Gram matrix is defined as
\(
    (G^{\mathrm{lap}}_U)_{ij}
    = \exp\!\left( -\gamma \, \| u_i - u_j \|_{1} \right),
\)
where $\gamma > 0$ is a kernel coefficient and $\|\cdot\|_{1}$ denotes the $\ell_1$ distance.

\subsection{Mutual information estimation}

To estimate the mutual information between two random variables $U$ and $V$, we compute their Gram matrices $G_U$ and $G_V$, and form the joint similarity matrix via element-wise (Hadamard) product:
\(
G_{UV} = G_U \circ G_V.
\) After trace-normalization, mutual information is estimated by:
\(
I(U; V) = H(U) + H(V) - H(U, V),
\)
where $H(U, V) = -\text{tr}(G_{UV} \log G_{UV})$. The eigenvalue spectrum of $G_{UV}$ governs the joint entropy term; its shape reflects how much of the structure in $U$ and $V$ aligns. A more concentrated spectrum in $G_{UV}$ relative to $G_U$ and $G_V$ implies stronger dependence and thus higher mutual information.

\section{Dataset overview and statistics}
\label{appendix:dataset}

\paragraph{Data Statistics} We conduct experiments on four datasets with varying levels of complexity and structure: CLUTRR~\citep{sinha2019clutrr}, ECQA~\citep{aggarwal2021explanations}, CNN/DailyMail~\citep{see-etal-2017-get,NIPS2015_afdec700} and a custom-designed Synthetic Arithmetic dataset. Table~\ref{app:tab_data} summarizes key dataset statistics.

\begin{table}[t]
\begin{center}
\resizebox{\columnwidth}{!}{
\begin{tabular}{lcccccc}
\hline
\textbf{Dataset} & \textbf{\#Train} &\textbf{Train Seq. Len} & \textbf{\#Val} &\textbf{Val Seq. Len}& \textbf{\#Test} &\textbf{Test Seq. Len}\\
\hline
\textbf{CLUTRR} & 9,074 & 30 & 2,020 & 29 & 1,146 & 70\\
\textbf{ECQA} & 7,598 & 21 & 1,090 & 21 & 2,194 & 21\\
\textbf{CNN/DailyMail} & - & - & - & - & 11,490 & 3,967\\
\textbf{Synthetic Arithmetic} & 10,000 & 9 & 2,000 & 9 & 2,000 & 9\\
\hline
\end{tabular}}
\end{center}
\caption{Dataset statistics.}
\label{app:tab_data}
\end{table}

\subsection{CLUTRR}

CLUTRR (Compositional Language Understanding and Text-based Relational Reasoning)~\citep{sinha2019clutrr} is a diagnostic benchmark for evaluating relational inference in language models. Each example contains a story describing family relations, and the task is to infer the missing relationship between two entities. The distribution shift stems from clause lengths that are absent in the training set but present during evaluation. We use the task split ``gen\_train23\_test2to10'', where the model is trained on clause lengths 2 and 3 and evaluated on lengths 2 through 10.

\subsection{ECQA}

ECQA (Explanations for CommonsenseQA)~\citep{aggarwal2021explanations} is a commonsense multiple-choice question-answering dataset, where each question is accompanied by 5 answer options.

\subsection{CNN/DailyMail}

CNN/DailyMail~\citep{see-etal-2017-get,NIPS2015_afdec700} is a summarization dataset containing news articles as written by journalists at CNN and the Daily Mail.

\subsection{Synthetic Arithmetic Dataset}

We construct a synthetic diagnostic dataset to disentangle task-relevant signal learning from spurious noise memorization in a controlled setting. Each sample consists of a sequence of 10 symbolic elements, where the signal component follows an arithmetic progression modulo \(K\), and the remainder of each element is independently corrupted with random noise. By varying the modulus \(K\), we systematically control task complexity and introduce structured shifts in the data distribution.

\paragraph{Synthetic Arithmetic dataset construction.}

At each position $t$, the signal value is computed as:
\[
s_t = (s_0 + t \cdot d) \bmod K, \quad \text{with } s_0 \in [0, K{-}1], \quad d \in [1, K{-}1].
\]
Each element in the sequence is represented as a string of the form:
\[
\texttt{S\{signal\}\_N\{noise\}}, \quad \text{where noise} \sim \mathcal{U}_\text{int}(0, \texttt{noise\_range-1}).
\]
Here $\mathcal{U}_\text{int}$ denotes the uniform distribution. The model is trained to predict the signal value of the final (10th) element, using the preceding elements as input context.

For example, with \(K = 5\), \(s_0 = 1\), and \(d = 2\), a sample might look like:
\begin{quote}
\resizebox{0.8\textwidth}{!}{\texttt{S1\_N42 S3\_N77 S0\_N18 S2\_N56 S4\_N90 S1\_N11 S3\_N65 S0\_N23 S2\_N37}}
\end{quote}
Each element encodes both a signal (the number following \texttt{S}) and a noise component (the number following \texttt{N}). The target is \texttt{4}, corresponding to the signal of the final (10th) item in the sequence.
\section{Implementation details}
\label{appendix:prompt}

This appendix outlines implementation details in our experiments.

\subsection{Prompt construction}

All tasks are cast into a natural language generation format. The model receives a prompt and generates the next token. Below are construction strategies and examples for each dataset:

\paragraph{CLUTRR}

Each input example in CLUTRR consists of a short narrative describing a set of family relationships, along with a query involving a pair of entities. We construct prompts by concatenating the narrative and a structured natural language question derived from the query tuple. The model is trained to predict the correct relationship as the next token.

\textbf{Prompt:}
\begin{quote}
Story: [Alice] is [Bob]'s mother. [Bob] is [Charlie]'s father. \\
Query: What is the relationship between Alice and Charlie? Answer:
\end{quote}

\textbf{Target:}~grandmother

\paragraph{ECQA (Explanation-augmented Commonsense QA)}

Each ECQA instance consists of a multiple-choice question with five candidate answers. We format the prompt by presenting the question followed by all five options (labeled A–E), and conclude with an explicit answer query. The model is trained to predict the correct answer letter as the next token.

\textbf{Prompt:}
\begin{quote}
Question: What do people usually do at a birthday party? \\
Options: \\
A. Sleep \\
B. Celebrate \\
C. Cook \\
D. Exercise \\
E. Drive \\
Answer:
\end{quote}

\textbf{Target:}~B

\paragraph{CNN / DailyMail}

Each input example consists of an article and its corresponding highlights.
We construct the prompt by appending the string
\texttt{\textbackslash n\textbackslash n The following is a one-paragraph highlight summary of the article:\textbackslash n\textbackslash n}
to the end of the article text.

\textbf{Prompt:}
\begin{quote}
(CNN)The Palestinian Authority officially became the 123rd member of the International Criminal Court on ...
\\
\\
The following is a one-paragraph highlight summary of the article:
\\
\end{quote}

\textbf{Target:}~Membership gives the ICC jurisdiction over alleged crimes committed in Palestinian territories since last June . Israel and the United States opposed the move, which could open the door to war crimes investigations against Israelis .

\paragraph{Synthetic Arithmetic}

Each synthetic sample consists of a sequence of 10 symbolic elements, where each element is formatted as \texttt{S\{signal\}\_N\{noise\}}. The signal values follow an arithmetic progression modulo $K$, and the noise values are independently sampled from a uniform distribution with a fixed range of 100. During training, the modulus $K$ is set to 13. For evaluation, test sequences are generated using values of $K$ from the range $[5, 26]$ excluding 13 to simulate a distribution shift. The model receives the first 9 tokens as input and is trained to predict the signal component of the 10th token.

\textbf{Prompt:}
\begin{quote}
S1\_N42 S3\_N88 S5\_N20 S7\_N10 S9\_N65 S11\_N43 S0\_N99 S2\_N38 S4\_N77
\\
\end{quote}

\textbf{Target:}~6

This controlled format enables manipulation of distributional properties by varying the modulus \(K\).

\subsection{Residual scaling with learnable \(\beta_\ell\) parameters}

We introduce a vector of learnable scalar weights \(\beta = \{\beta_1, \dots, \beta_L\}\) applied to residual connections in a frozen transformer:

\[
z^{(\ell)} = z^{(\ell-1)} + \beta_\ell \cdot \text{Block}^{(\ell)}(z^{(\ell-1)}).
\]

\begin{itemize}
  \item All transformer weights are frozen; only \(\beta_\ell\) parameters are trained.
  \item Each transformer block's residual output is modulated by \(\beta_\ell\) via forward hooks.
  \item \(\beta\) are initialized to 1 and updated using gradient descent without altering the architecture.
\end{itemize}

For Synthetic Arithmetic, the in-distribution (ID) setting is using $K = 13$, the out-of-distribution (OOD) setting is using $K = [5, 26]$ excluding 13. For CLUTRR, the in-distribution (ID) setting is using the ``1.2'' and ``1.3'' split of ``gen\_train23\_test2to10''. Out-of-distribution (OOD) is conducted on held-out CLUTRR configurations that do not overlap with the training distribution: the ``1.4'' to ``1.10'' split of ``gen\_train23\_test2to10''. For ECQA, we use training set as the ID setting and CLUTRR as the OOD setting. The ID and OOD sizes are balanced.

\subsection{Training and evaluation settings}

\begin{wraptable}{r}{0.5\textwidth}
\begin{center}
\label{tab:results}
\resizebox{0.5\textwidth}{!}{%
\begin{tabular}{lccc}
\toprule
\textbf{Model} & \textbf{CLUTRR} & \textbf{Synthetic} & \textbf{ECQA} \\
\midrule
GPT-2 Small     & 56.57 & 72.55 & --- \\
GPT-2 Medium    & 62.11 & 74.05 & --- \\
Qwen-2.5-0.5B   & 58.61 & 88.00 & 65.45 \\
LLaMA-3.1-8B    & 72.62 & 89.70 & 80.49 \\
\bottomrule
\end{tabular}
}
\end{center}
\caption{Performance (\%) of different models across datasets.}
\end{wraptable}

We finetune the model using HuggingFace’s Trainer API with a batch size of 16, a learning rate of $\alpha = 5 \times 10^{-6}$, and AdamW optimizer with weight decay of 0.01. We perform residual $\beta_\ell$ training on top of a model that has already been fine-tuned on the target dataset, keeping all previously learned weights frozen and optimizing only the $\beta_\ell$ parameters. We use a batch size of 4, a learning rate of 0.001, Adam optimizer and epochs of 10. Evaluation is conducted by comparing the next predicted token with the next ground-truth target token.
\section{Compute and licensing details}
\label{appendix:compute}

\paragraph{Computing resources} Experiments were conducted on an NVIDIA RTX A6000 GPU (48GB).

\paragraph{Model licenses} The GPT-2 Small~\citep{radford2019language} and GPT-2 Medium~\citep{radford2019language} models are released under the MIT License and are available via Hugging Face Transformers. The Qwen-2.5 0.5B~\citep{qwen2} model is provided by Alibaba Group under the Apache 2.0 License. The LLaMA-3.1 8B~\citep{meta_llama_3.1_8b} model is made available by Meta under a non-commercial research license and accessed via Hugging Face.

\paragraph{Dataset licenses} CLUTRR~\citep{sinha2019clutrr} is released under a CC BY 4.0 license as part of EMNLP 2019. The ECQA~\citep{aggarwal2021explanations} dataset is also released under a CC BY 4.0 license by its authors as part of EMNLP 2021. CNN/DailyMail is under Apache-2.0 License. The Synthetic Arithmetic dataset is custom-designed by the authors and does not rely on any external or licensed data sources.

All assets were used in compliance with their respective licenses, and no proprietary or restricted resources were employed in our experiments.

\newpage
\section{Full quantitative results with confidence estimates}
\label{appendix:fullresults}

\paragraph{Predictive information}

To assess estimation variability and sensitivity with the choice of evaluation data, we report the results with statistical significance over 5 random 100 subsamples of test examples.

\begin{figure}[H]
    \centering
    \begin{subfigure}[b]{0.4\textwidth}
        \includegraphics[width=\linewidth]{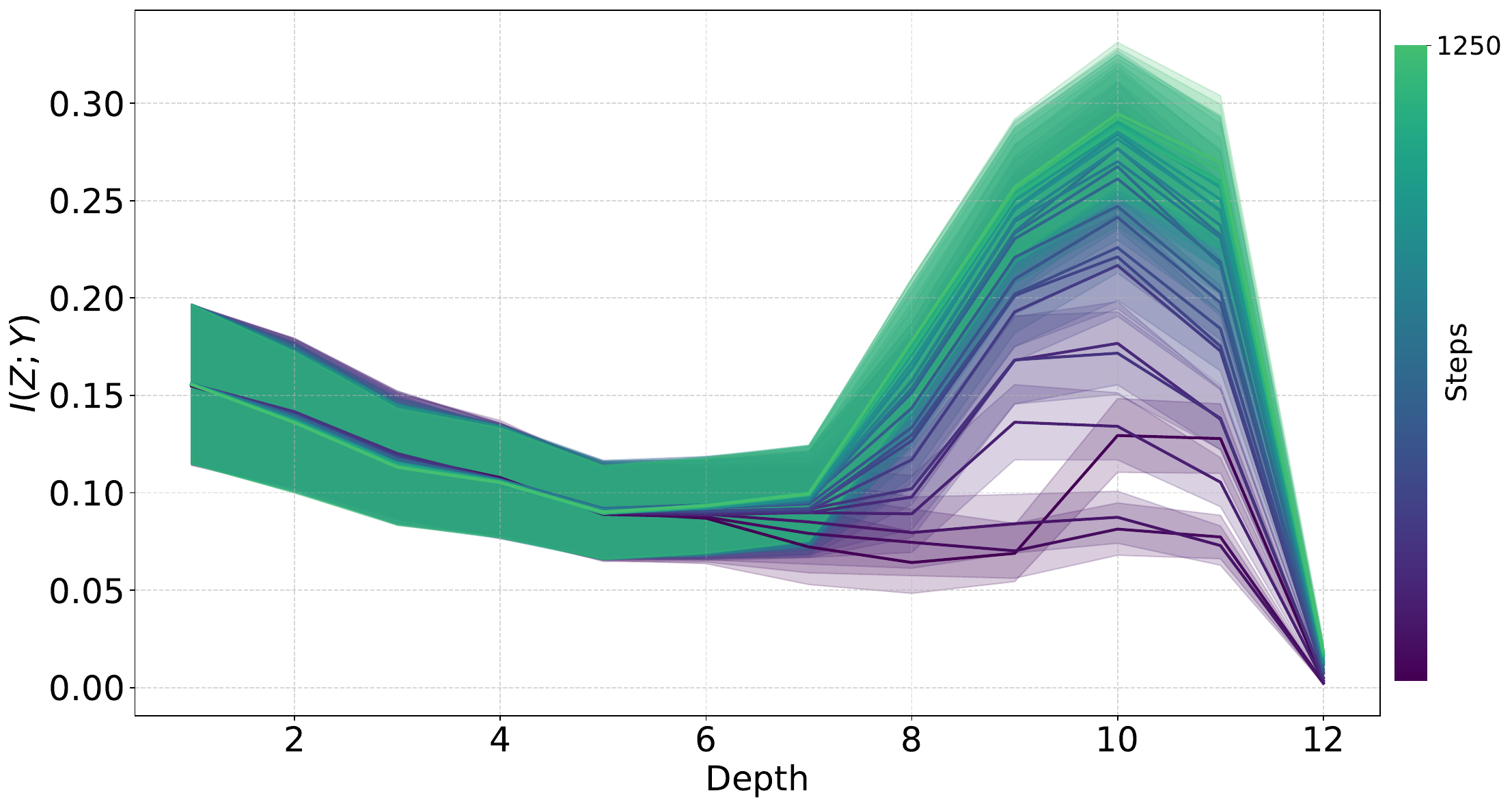}
        \caption{GPT-2 Small on CLUTRR}
    \end{subfigure}
    \hfill
    \begin{subfigure}[b]{0.4\textwidth}
        \includegraphics[width=\linewidth]{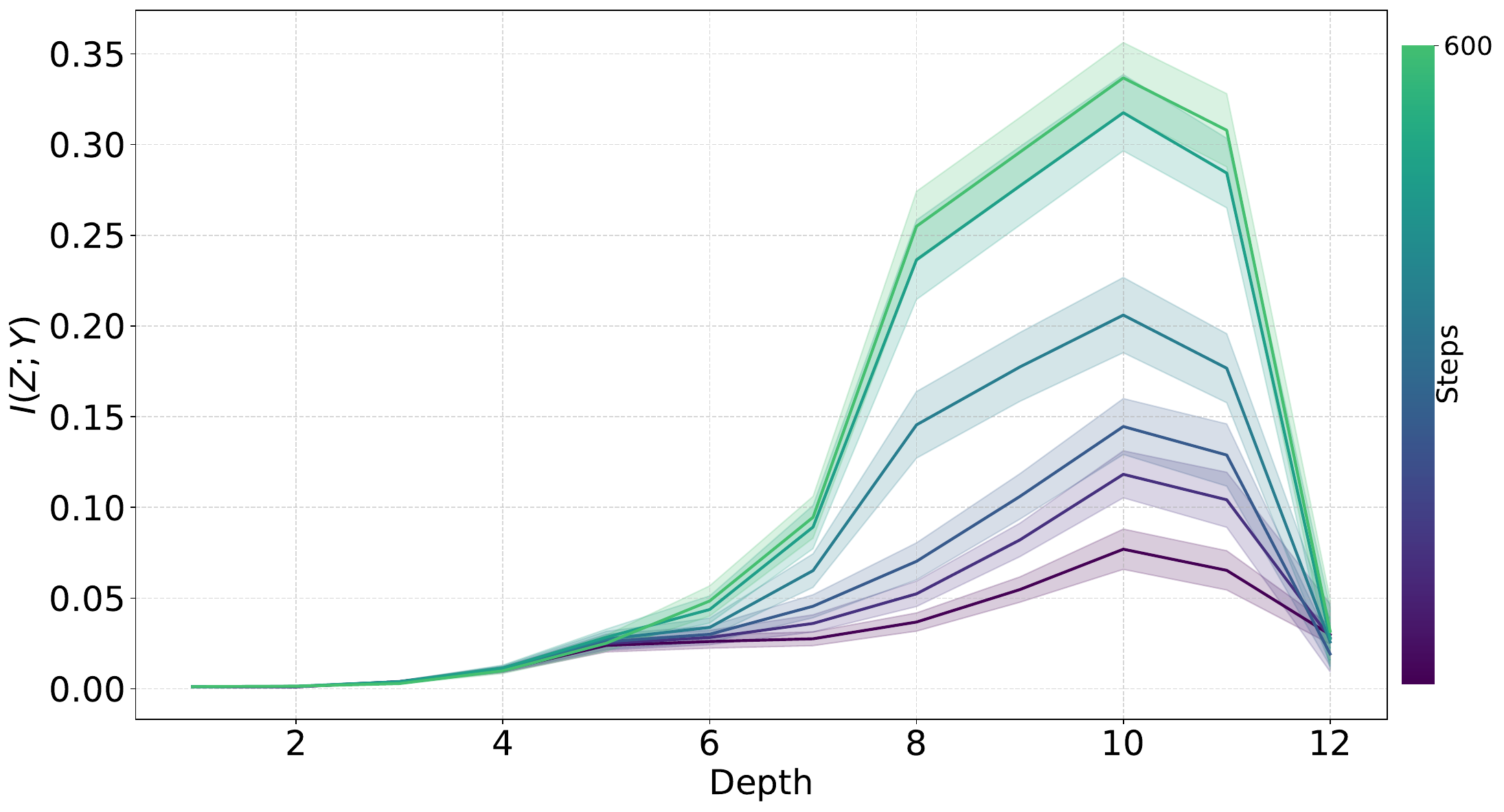}
        \caption{GPT-2 Small on Synthetic}
    \end{subfigure}
    \hfill
    \begin{subfigure}[b]{0.4\textwidth}
        \includegraphics[width=\linewidth]{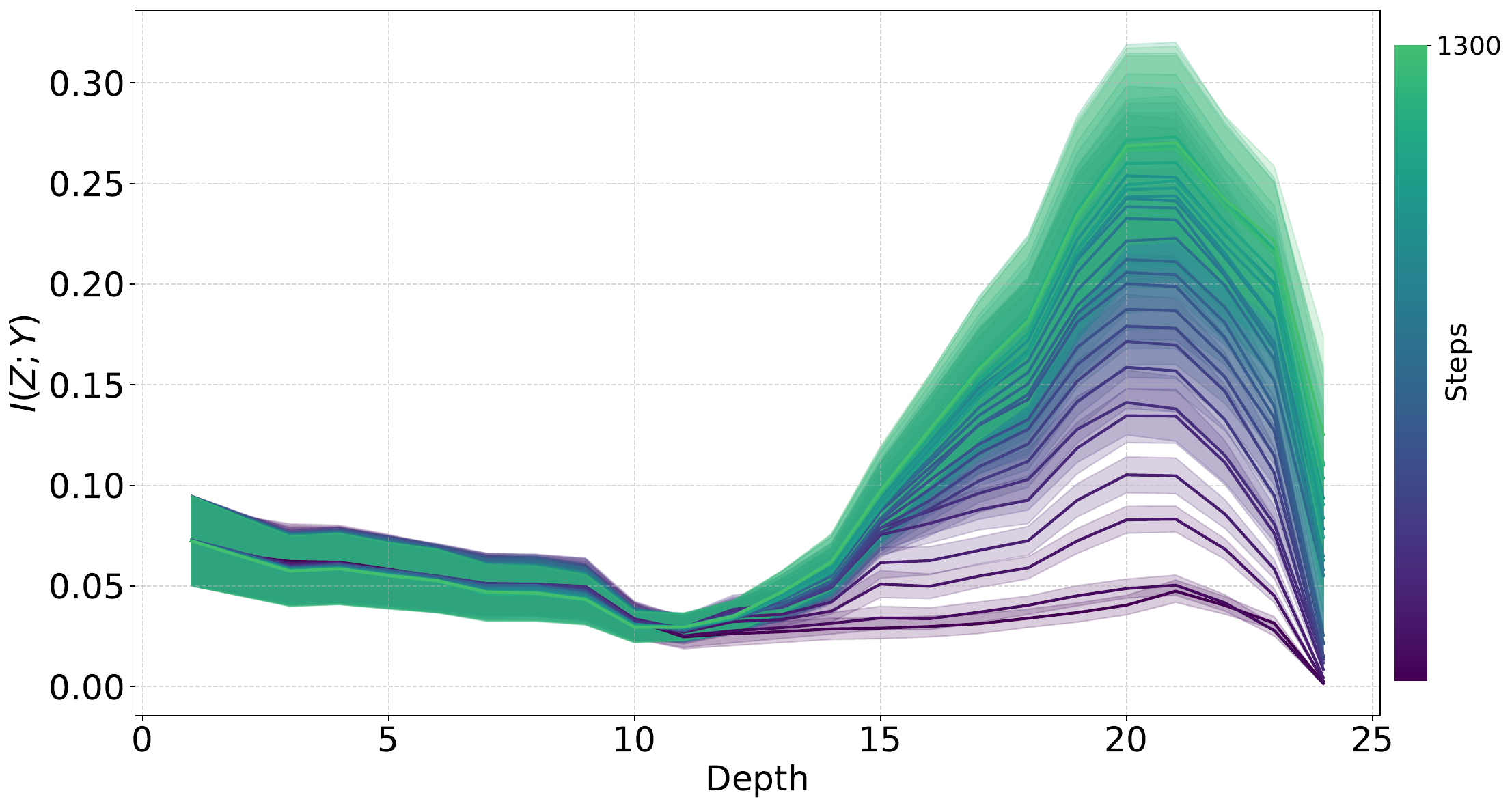}
        \caption{GPT-2 Medium on CLUTRR}
    \end{subfigure}
    \hfill
    \begin{subfigure}[b]{0.4\textwidth}
        \includegraphics[width=\linewidth]{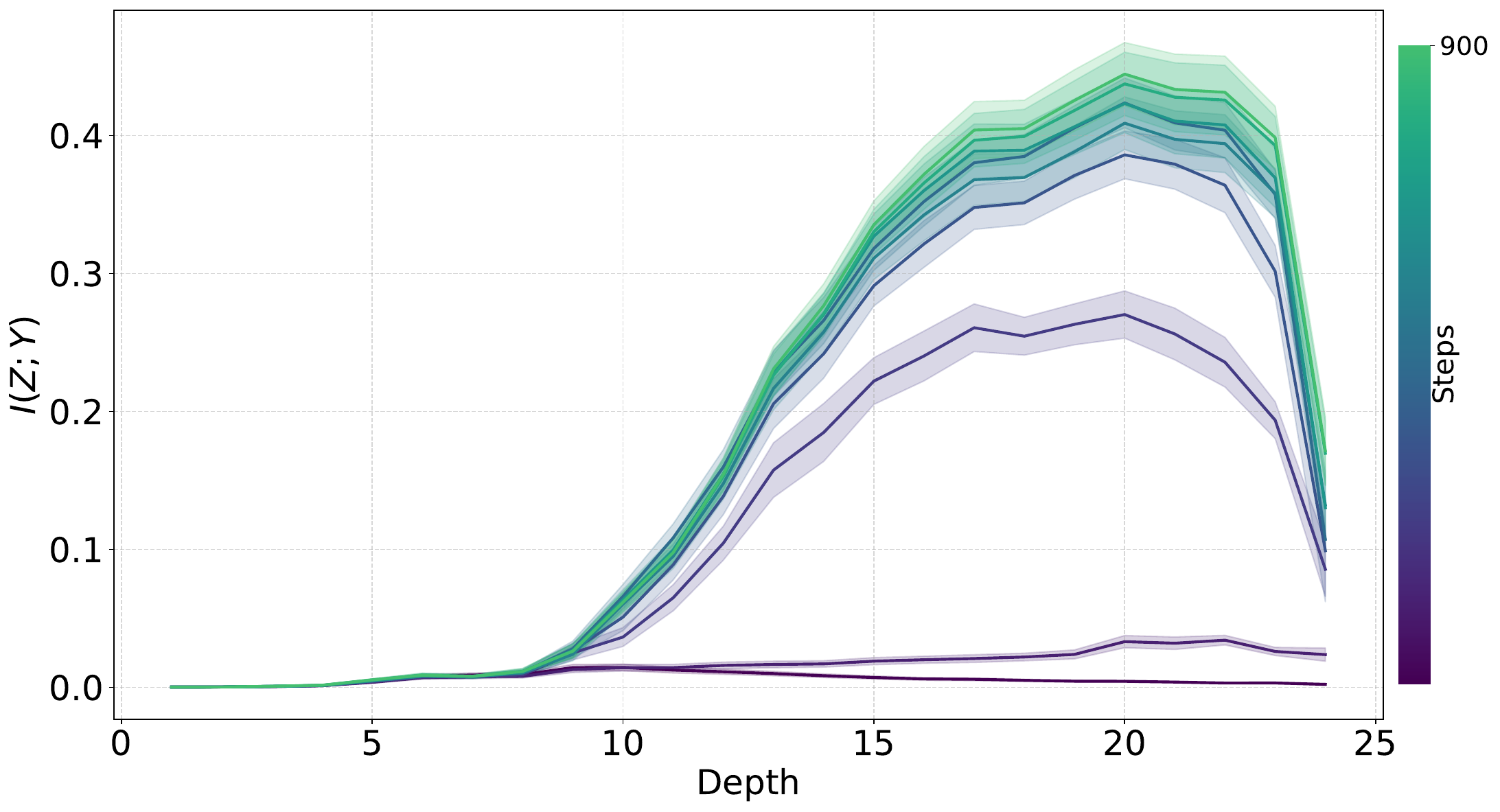}
        \caption{GPT-2 Medium on Synthetic}
    \end{subfigure}
    \hfill
    \begin{subfigure}[b]{0.4\textwidth}
        \includegraphics[width=\linewidth]{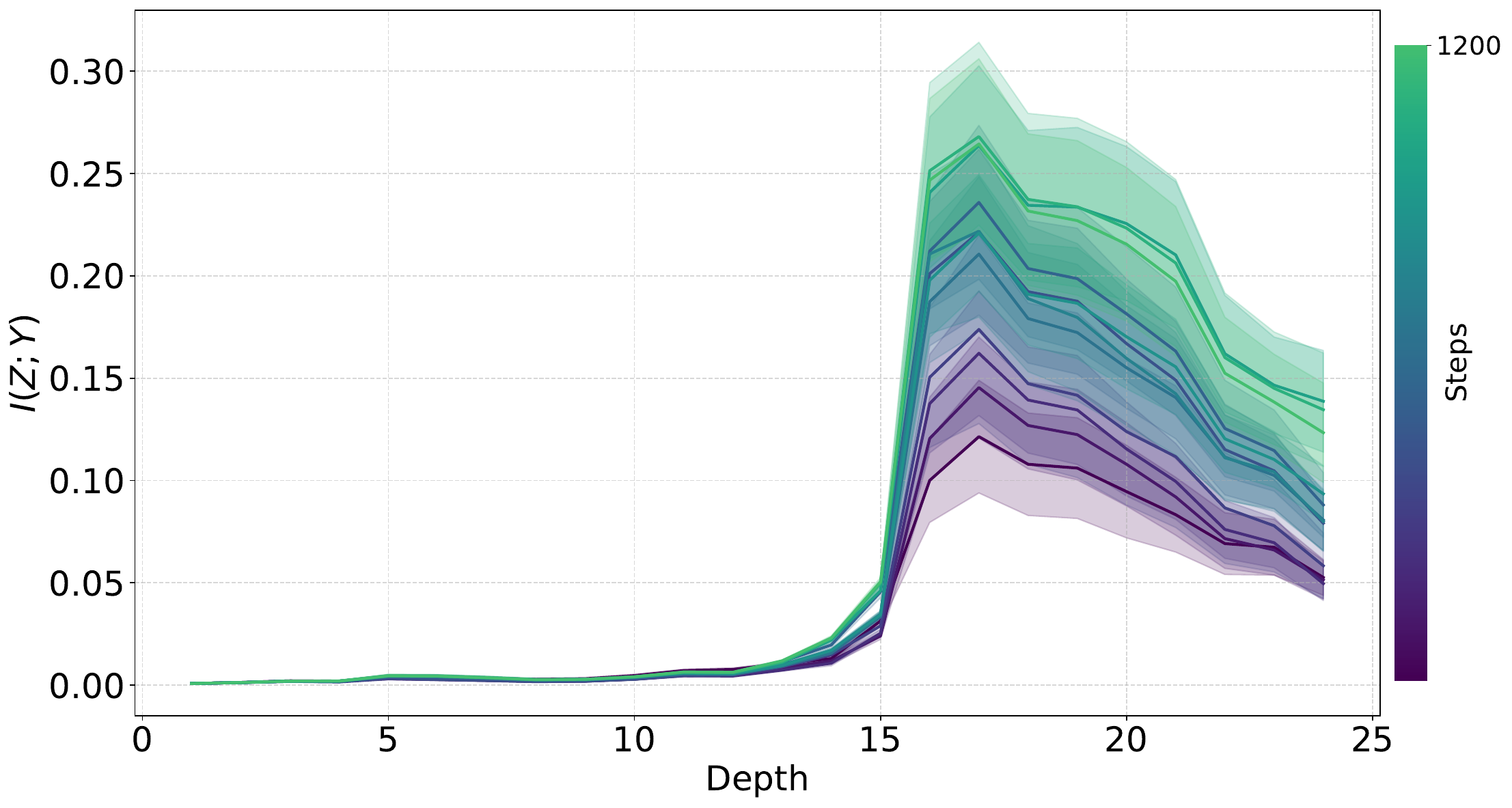}
        \caption{Qwen-2.5-0.5B on ECQA}
    \end{subfigure}
    \hfill
    \begin{subfigure}[b]{0.4\textwidth}
        \includegraphics[width=\linewidth]{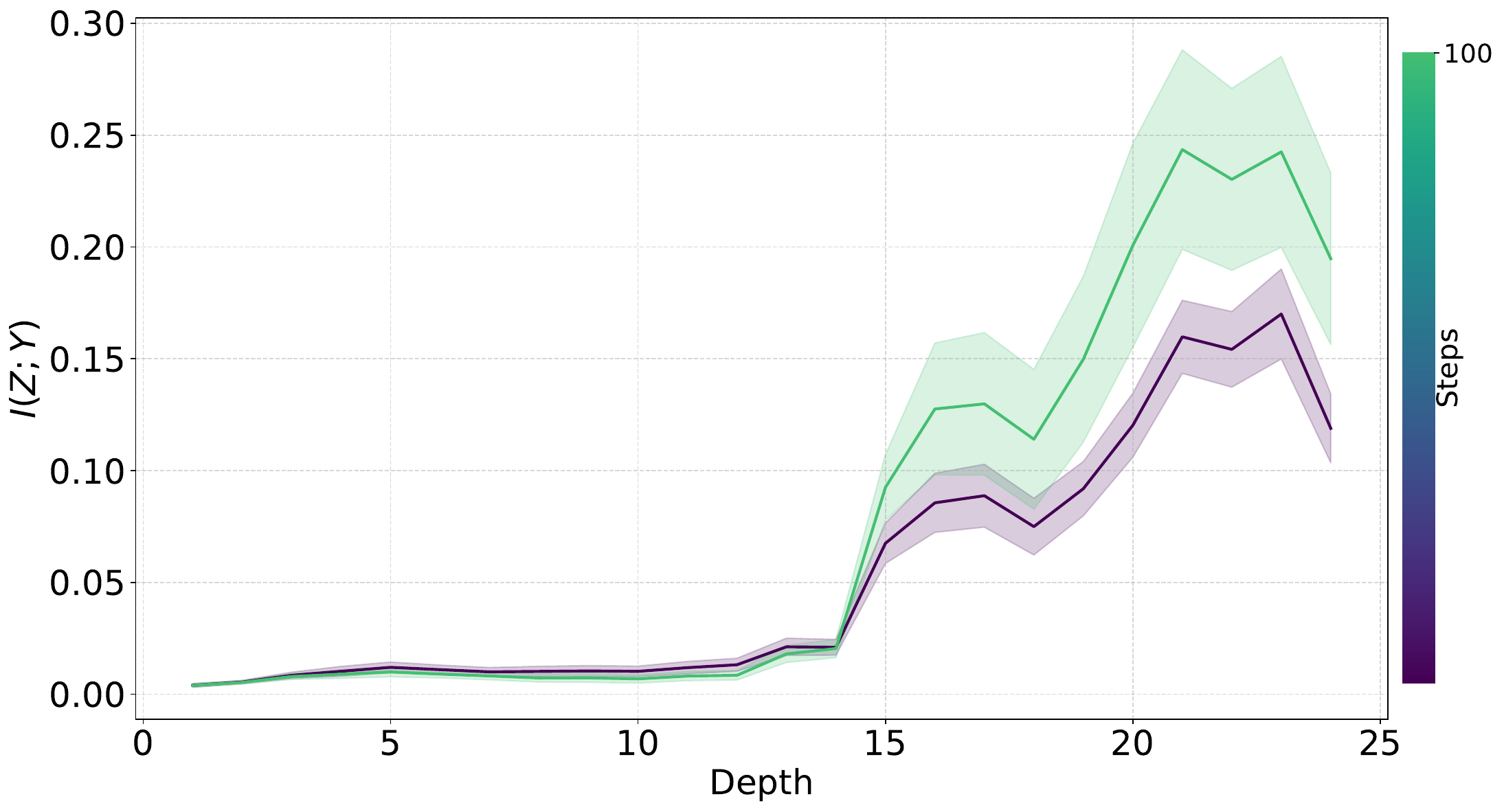}
        \caption{Qwen-2.5-0.5B on CLUTRR}
    \end{subfigure}
    \hfill
    \begin{subfigure}[b]{0.4\textwidth}
        \includegraphics[width=\linewidth]{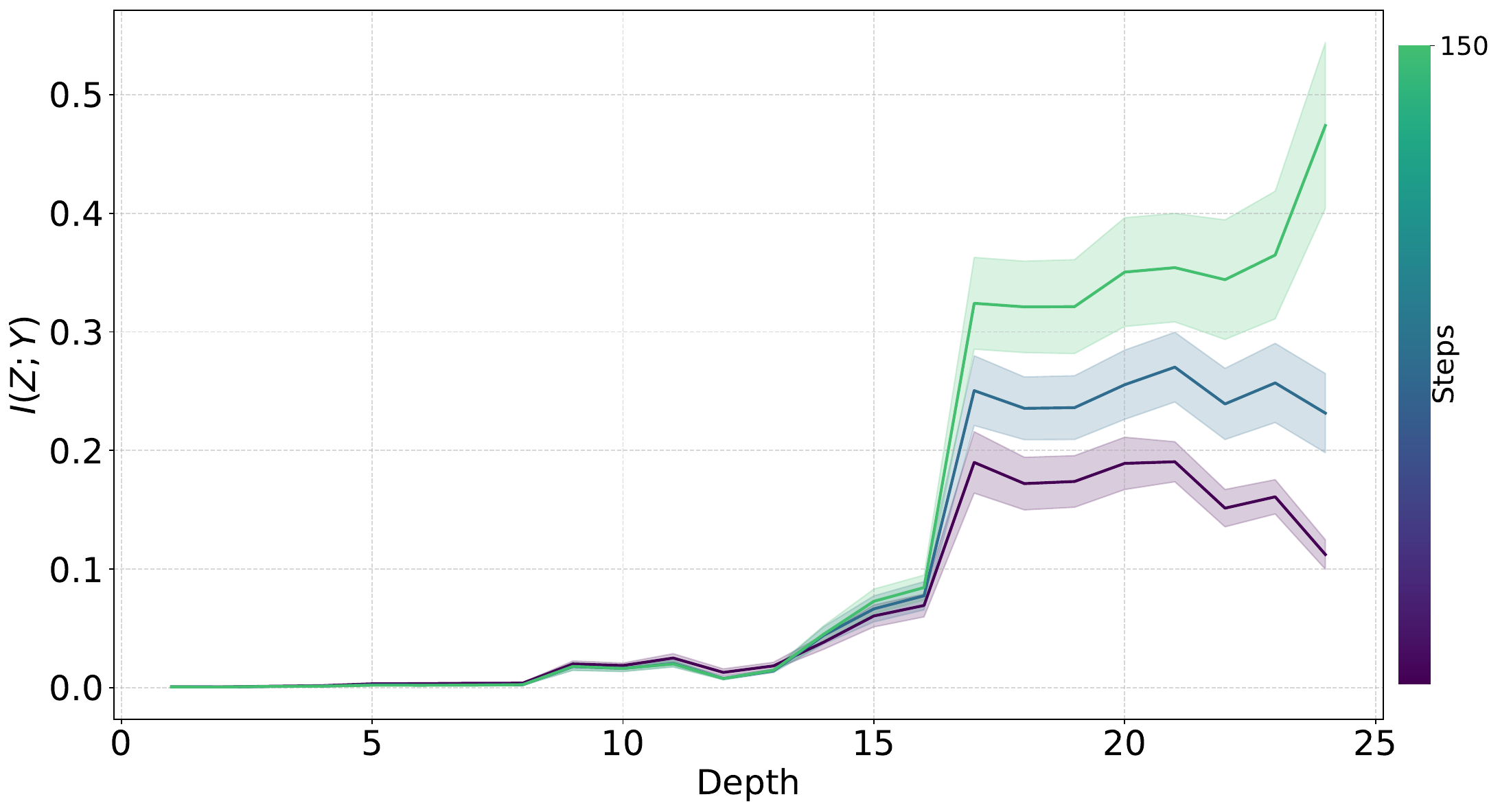}
        \caption{Qwen-2.5-0.5B on Synthetic}
    \end{subfigure}
    \hfill
    \begin{subfigure}[b]{0.4\textwidth}
        \includegraphics[width=\linewidth]{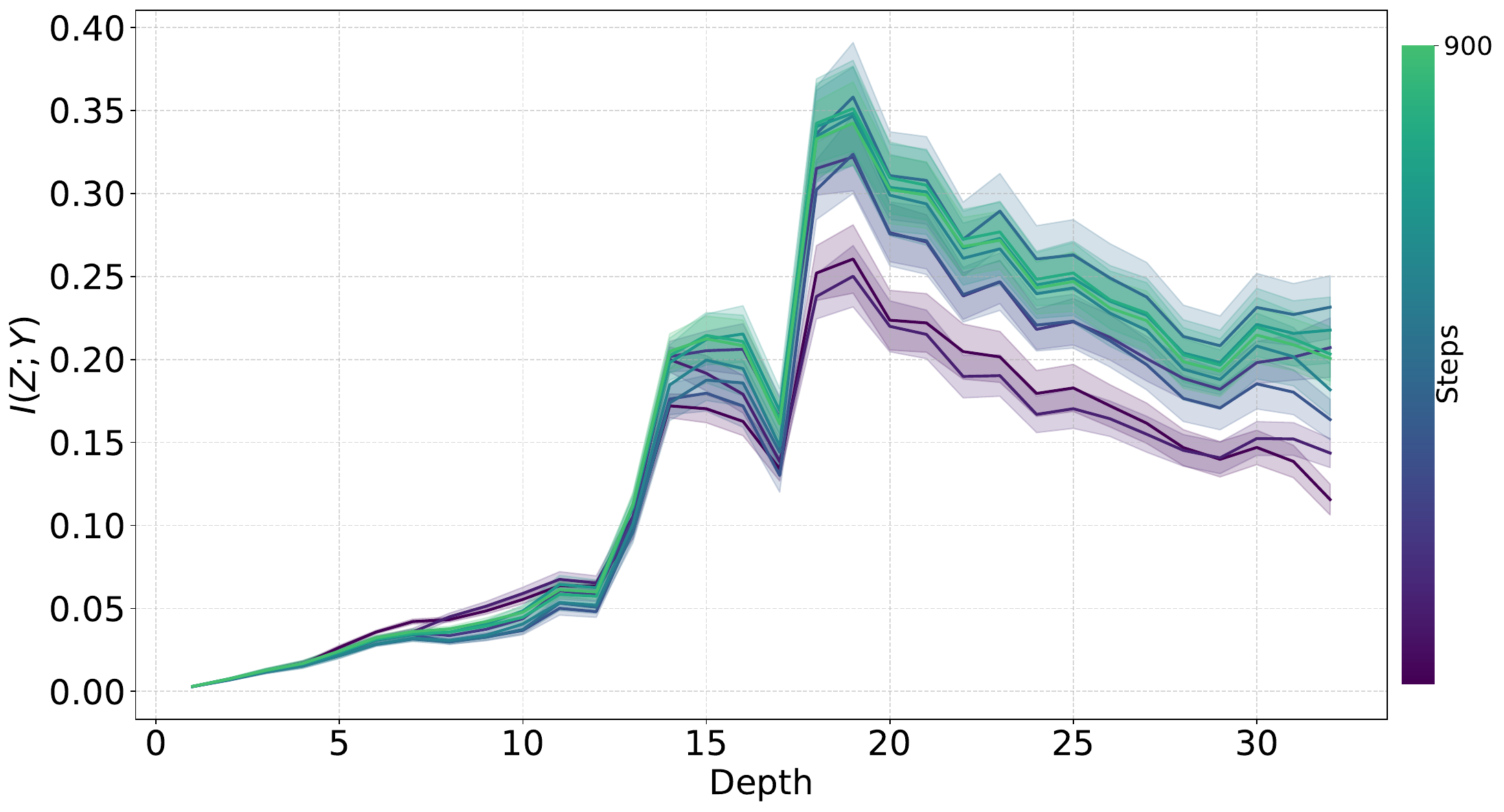}
        \caption{LLaMA-3.1-8B on ECQA}
    \end{subfigure}
    \hfill
    \begin{subfigure}[b]{0.4\textwidth}
        \includegraphics[width=\linewidth]{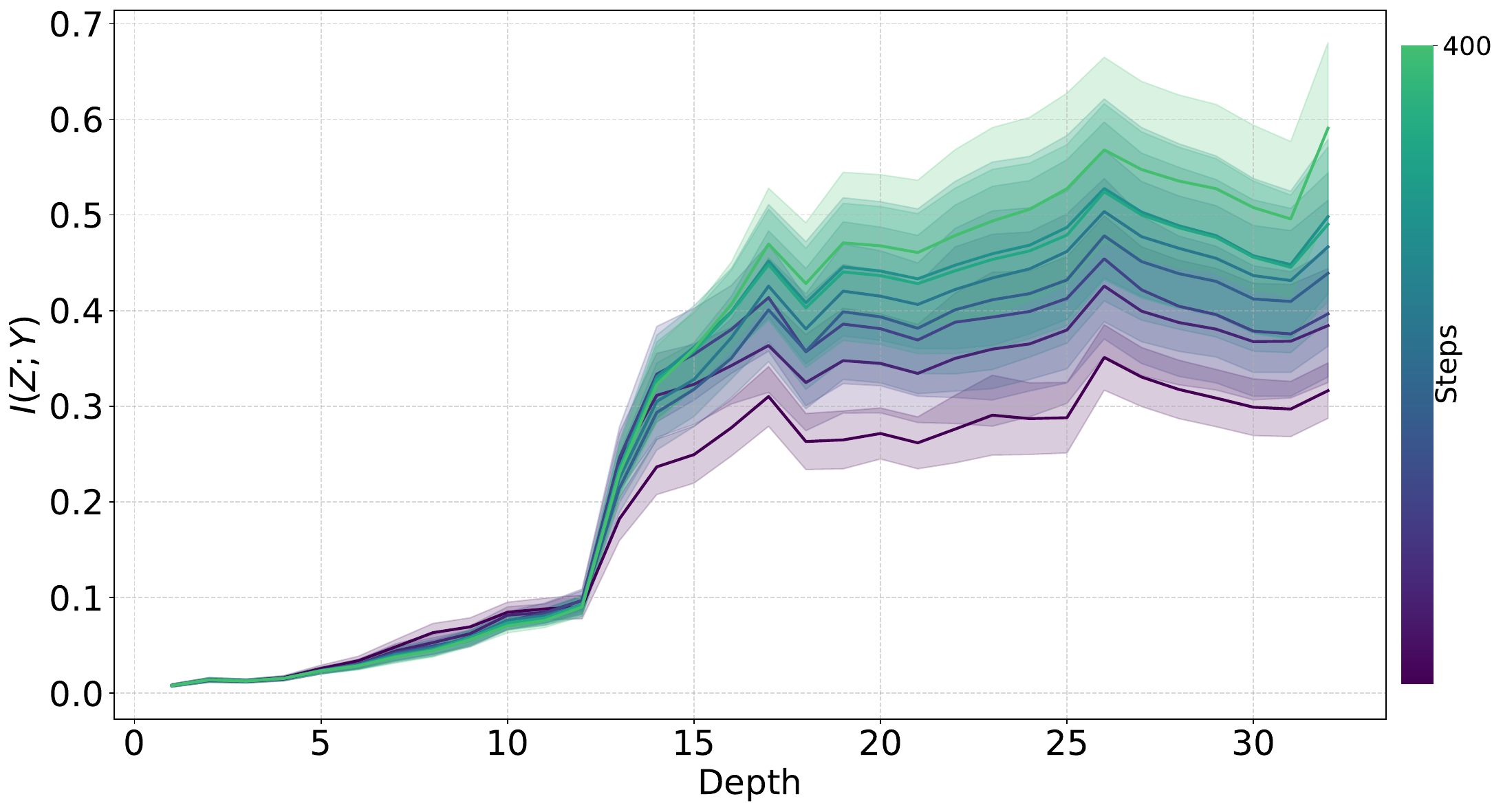}
        \caption{LLaMA-3.1-8B on CLUTRR}
    \end{subfigure}
    \hfill
    \begin{subfigure}[b]{0.4\textwidth}
        \includegraphics[width=\linewidth]{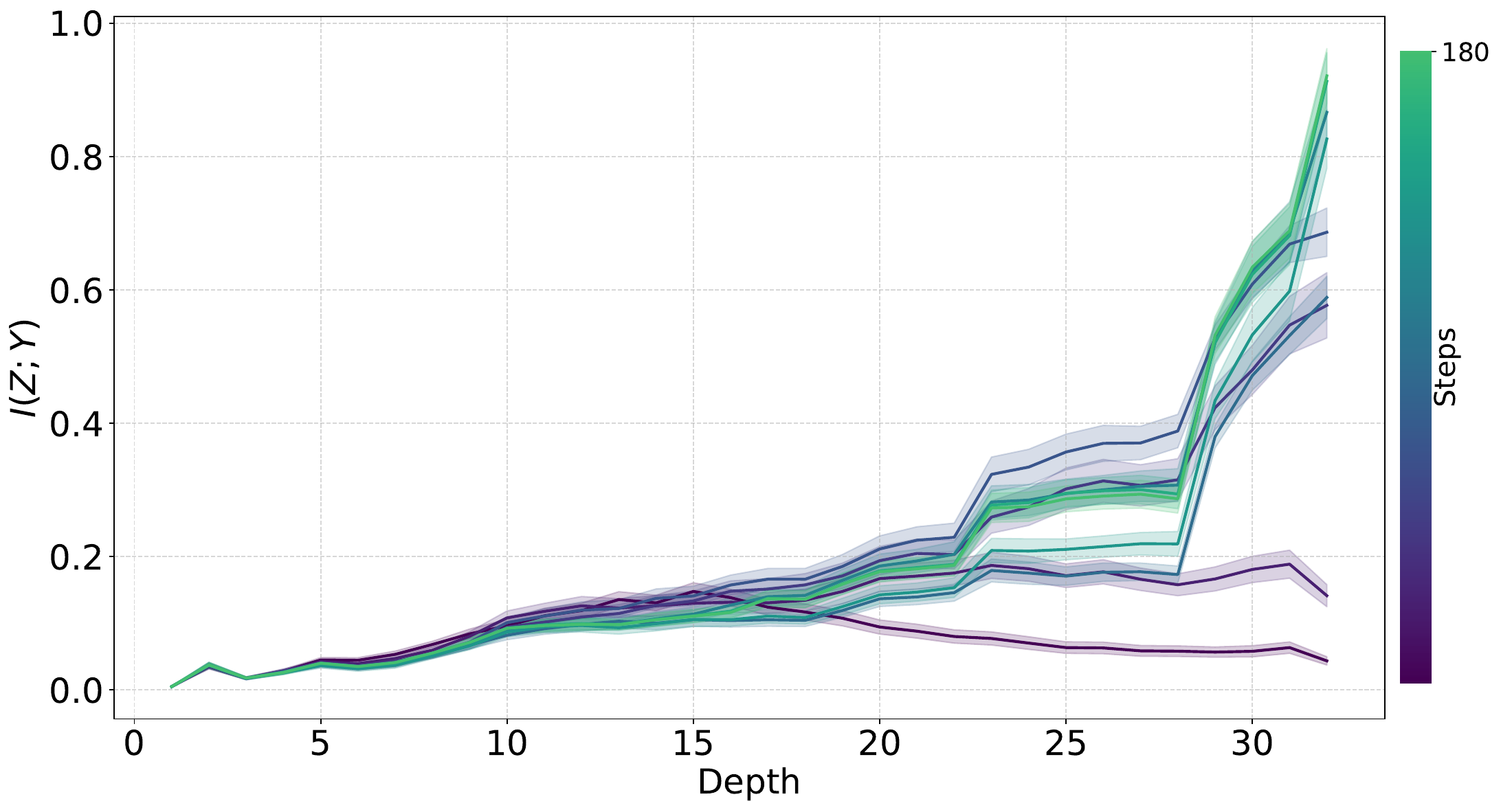}
        \caption{LLaMA-3.1-8B on Synthetic}
    \end{subfigure}
    \caption{Predictive information $I(Z;Y)$ across models and datasets. Lighter line colors represent later training steps. Each curve shows the mean across 5 sampling seeds (0, 1, 2, 3, 42), and the shaded region denotes a 2-sigma (\textasciitilde96\%) confidence interval.}
    \label{fig:app_izy_full}
\end{figure}

\paragraph{Incremental information gain}

In addition to predictive information \( I(Z^{(\ell)}; Y) \), we compute the incremental information gain \( I(\Delta Z^{(\ell)}; Y) \) for each transformer block.

\begin{figure}[H]
    \centering
    \begin{subfigure}[b]{0.47\textwidth}
        \includegraphics[width=\linewidth]{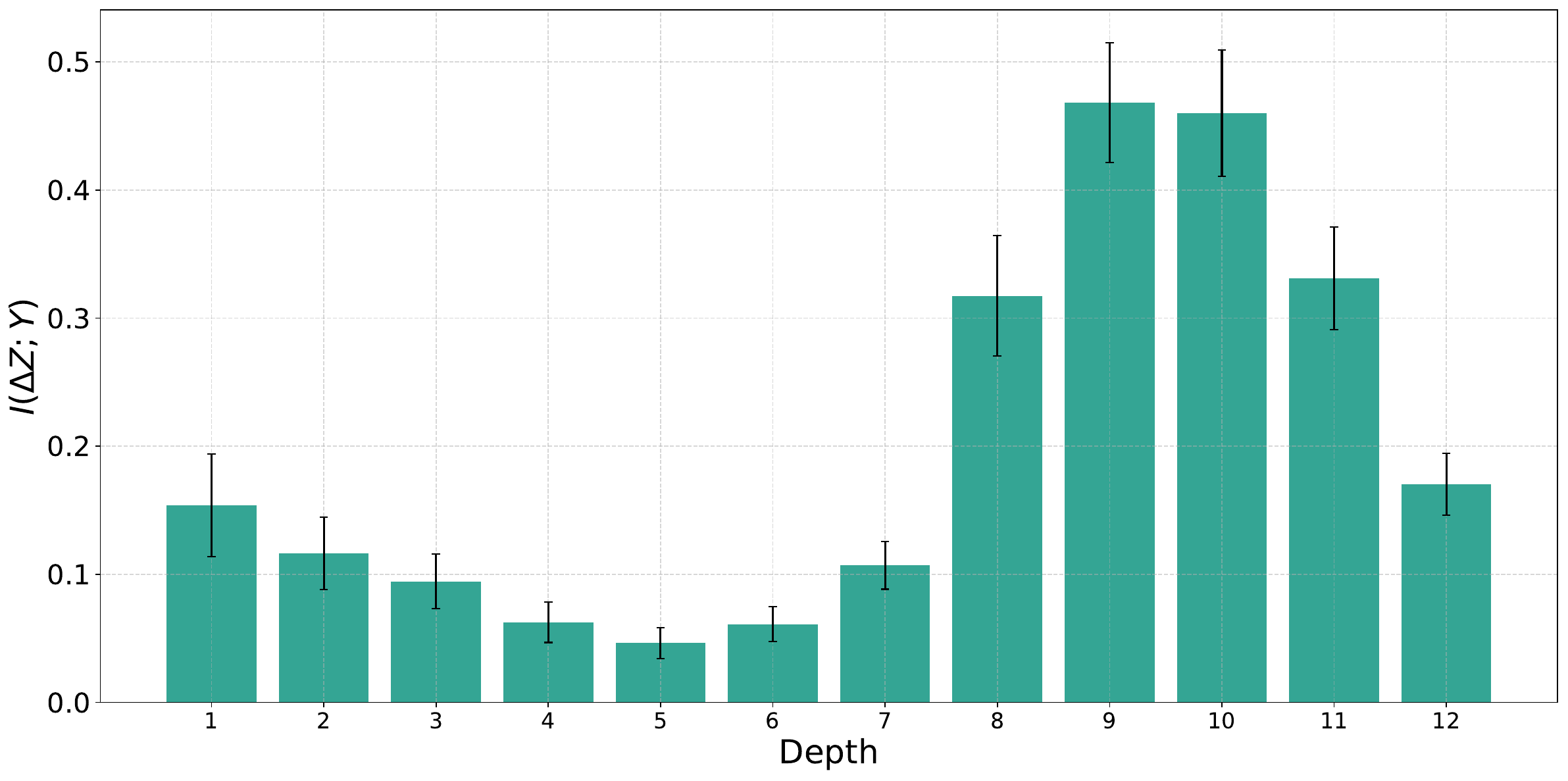}
        \caption{GPT-2 Small on CLUTRR}
    \end{subfigure}
    \hfill
    \begin{subfigure}[b]{0.47\textwidth}
        \includegraphics[width=\linewidth]{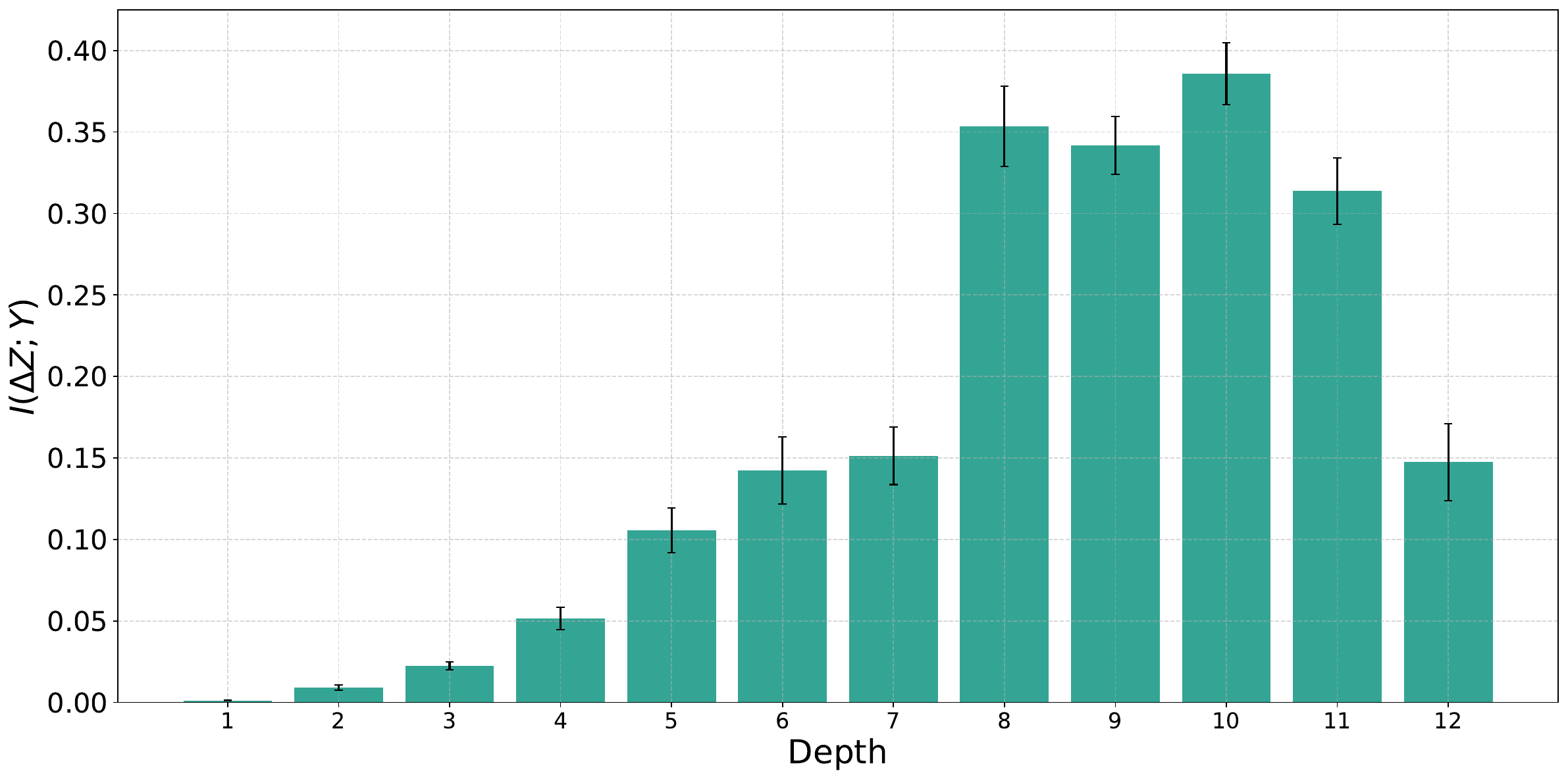}
        \caption{GPT-2 Small on Synthetic}
    \end{subfigure}
    \hfill
    \begin{subfigure}[b]{0.47\textwidth}
        \includegraphics[width=\linewidth]{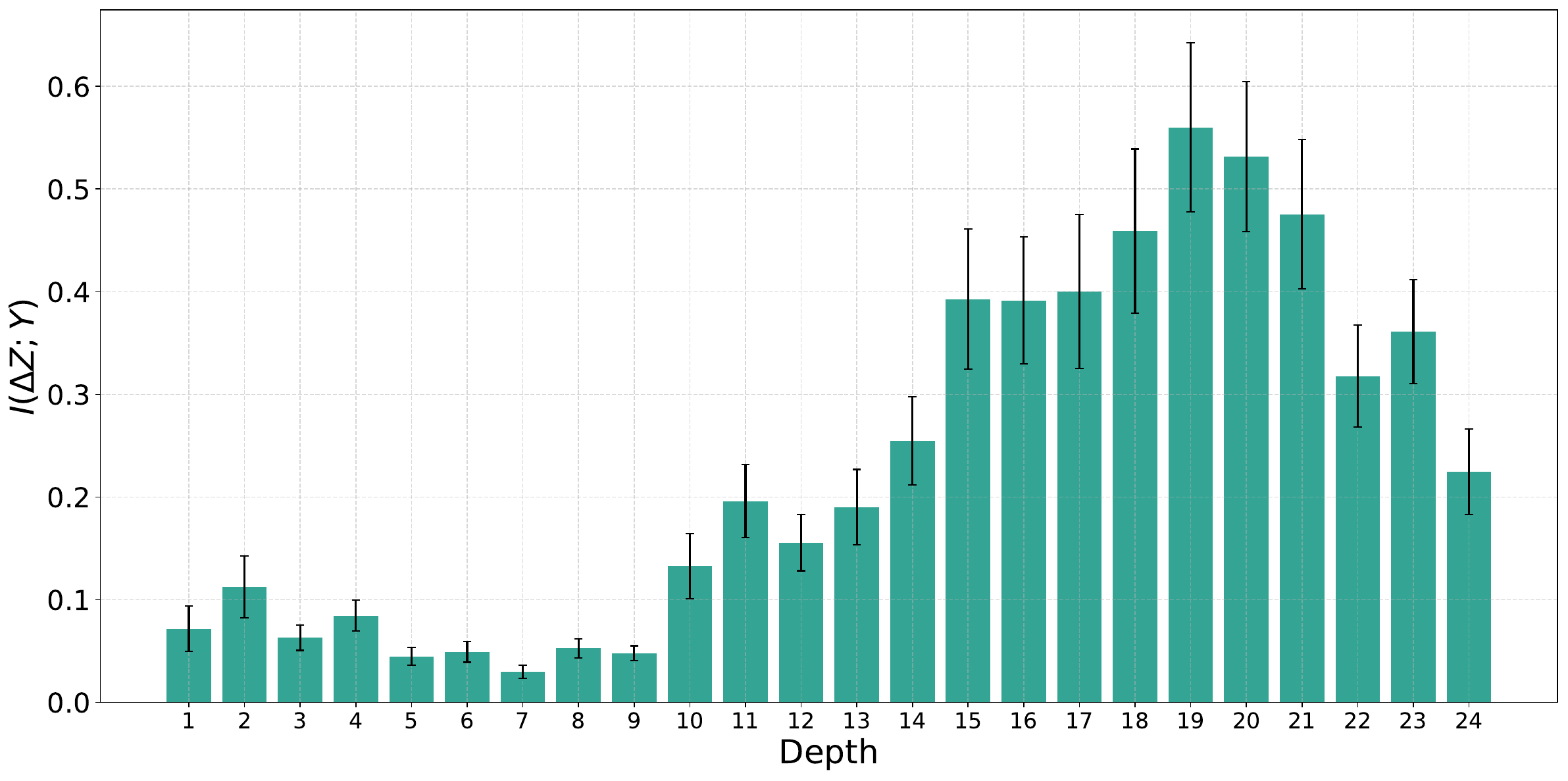}
        \caption{GPT-2 Medium on CLUTRR}
    \end{subfigure}
    \hfill
    \begin{subfigure}[b]{0.47\textwidth}
        \includegraphics[width=\linewidth]{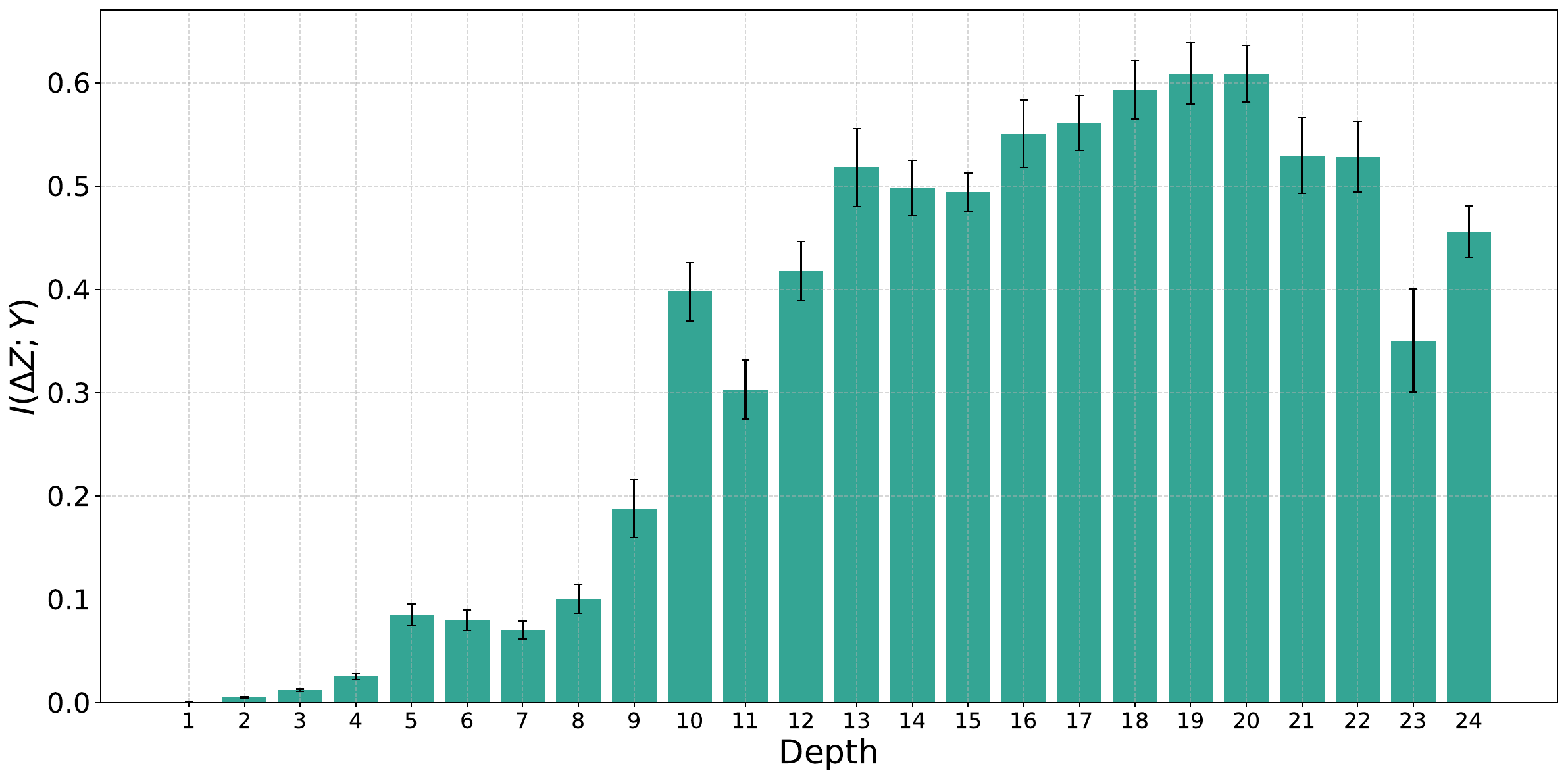}
        \caption{GPT-2 Medium on Synthetic}
    \end{subfigure}
    \hfill
    \begin{subfigure}[b]{0.47\textwidth}
        \includegraphics[width=\linewidth]{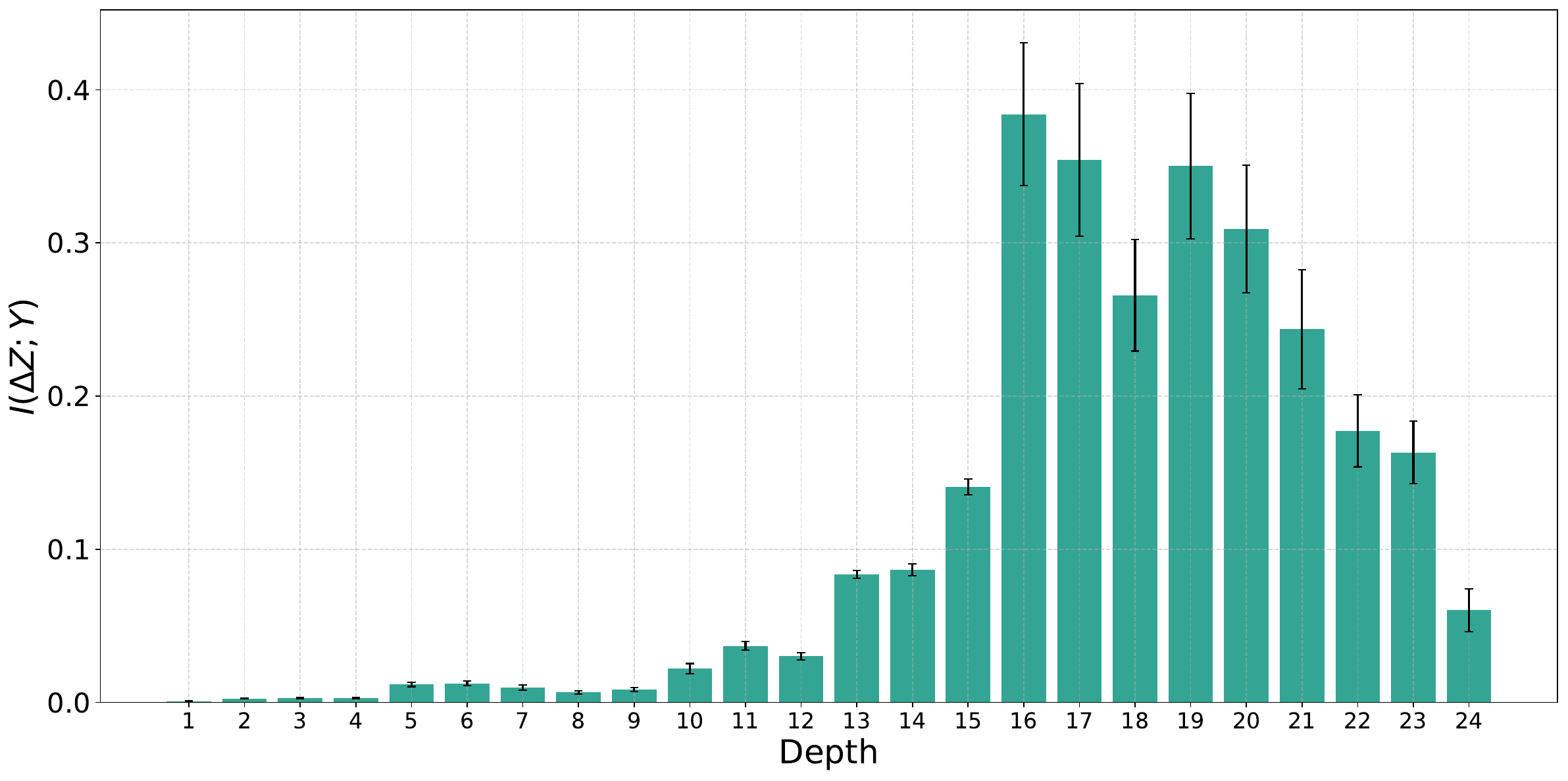}
        \caption{Qwen-2.5-0.5B on ECQA}
    \end{subfigure}
    \hfill
    \begin{subfigure}[b]{0.47\textwidth}
        \includegraphics[width=\linewidth]{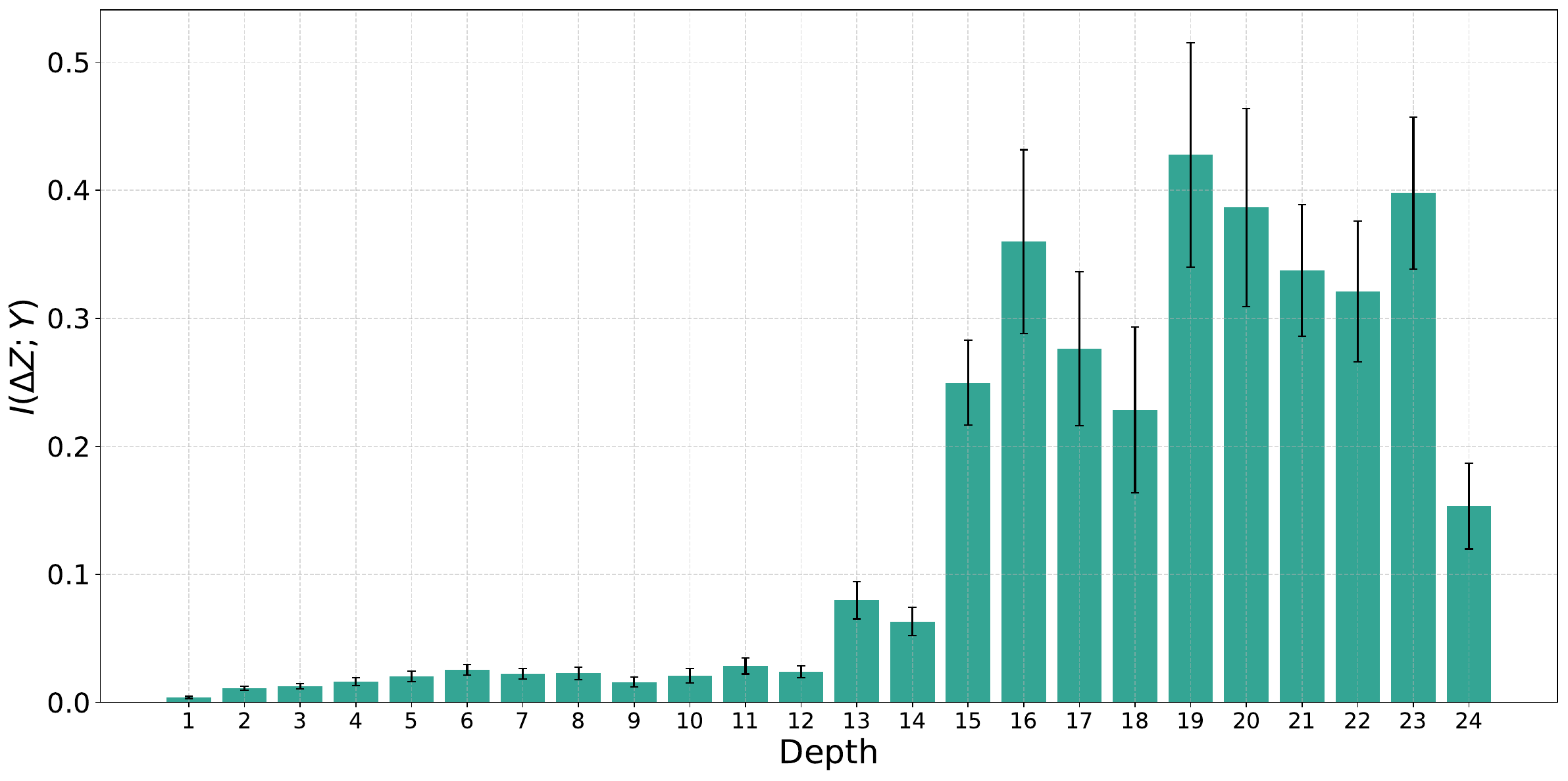}
        \caption{Qwen-2.5-0.5B on CLUTRR}
    \end{subfigure}
    \hfill
    \begin{subfigure}[b]{0.47\textwidth}
        \includegraphics[width=\linewidth]{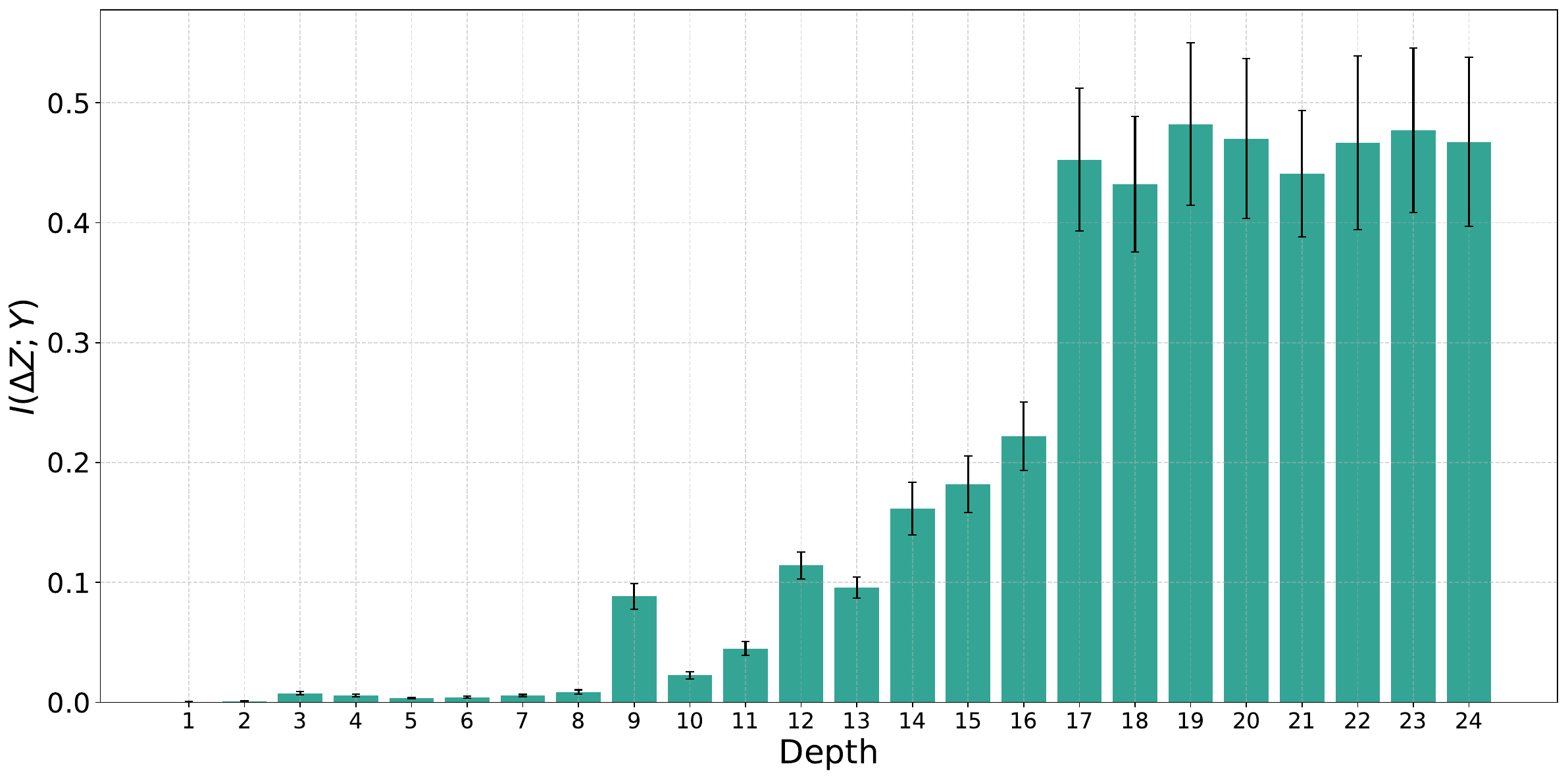}
        \caption{Qwen-2.5-0.5B on Synthetic}
    \end{subfigure}
    \hfill
    \begin{subfigure}[b]{0.47\textwidth}
        \includegraphics[width=\linewidth]{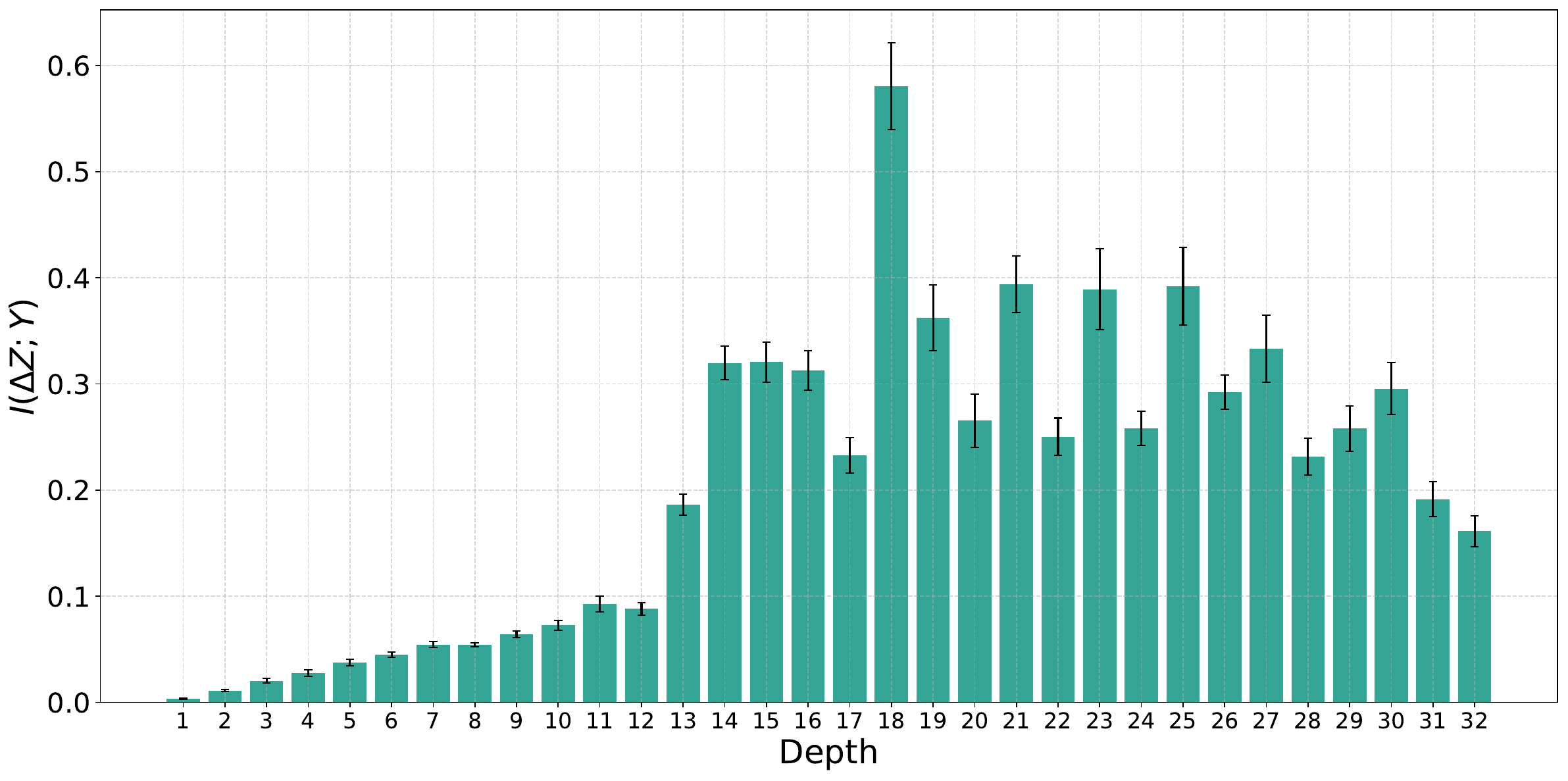}
        \caption{LLaMA-3.1-8B on ECQA}
    \end{subfigure}
    \hfill
    \begin{subfigure}[b]{0.47\textwidth}
        \includegraphics[width=\linewidth]{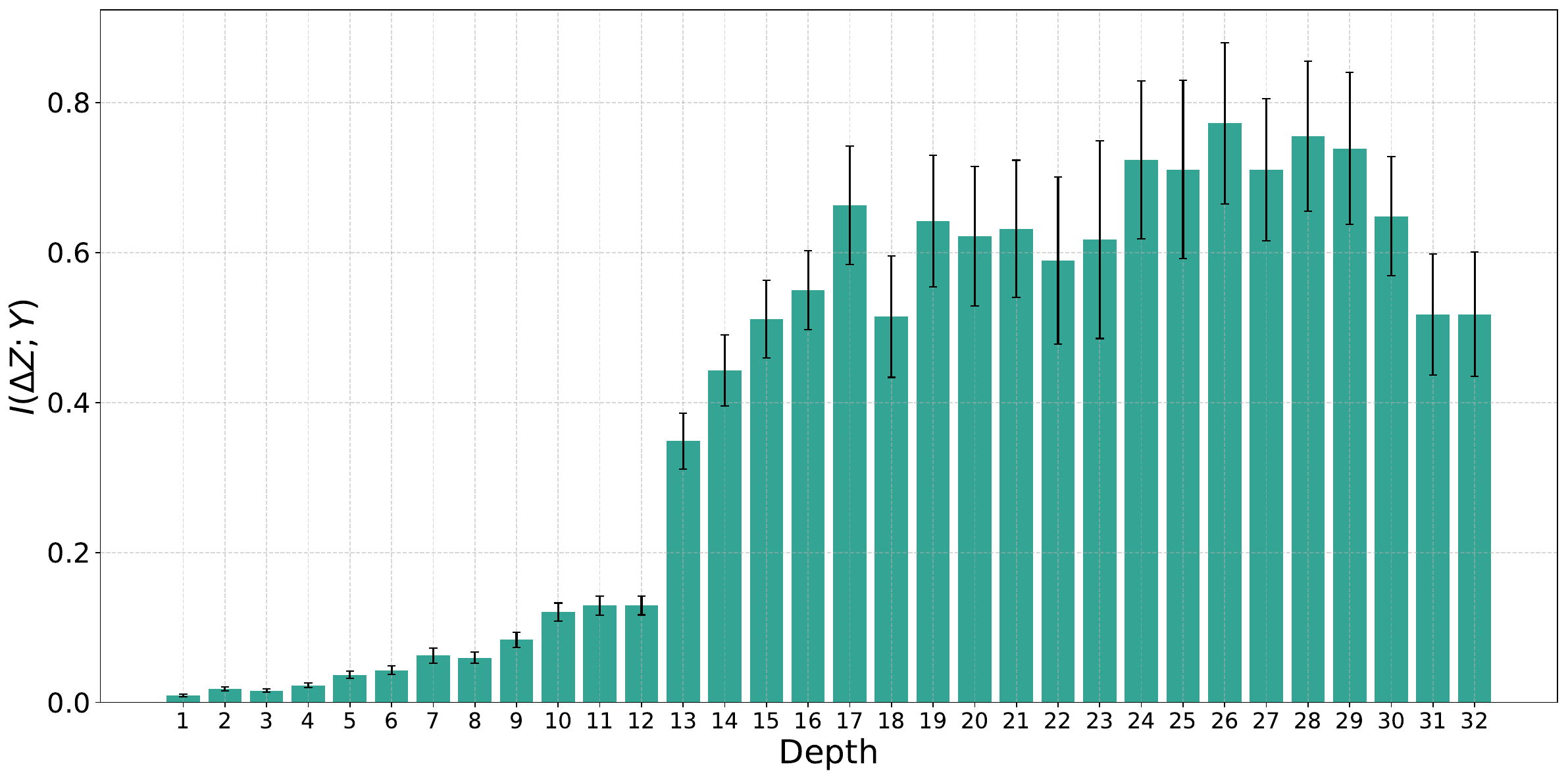}
        \caption{LLaMA-3.1-8B on CLUTRR}
    \end{subfigure}
    \hfill
    \begin{subfigure}[b]{0.47\textwidth}
        \includegraphics[width=\linewidth]{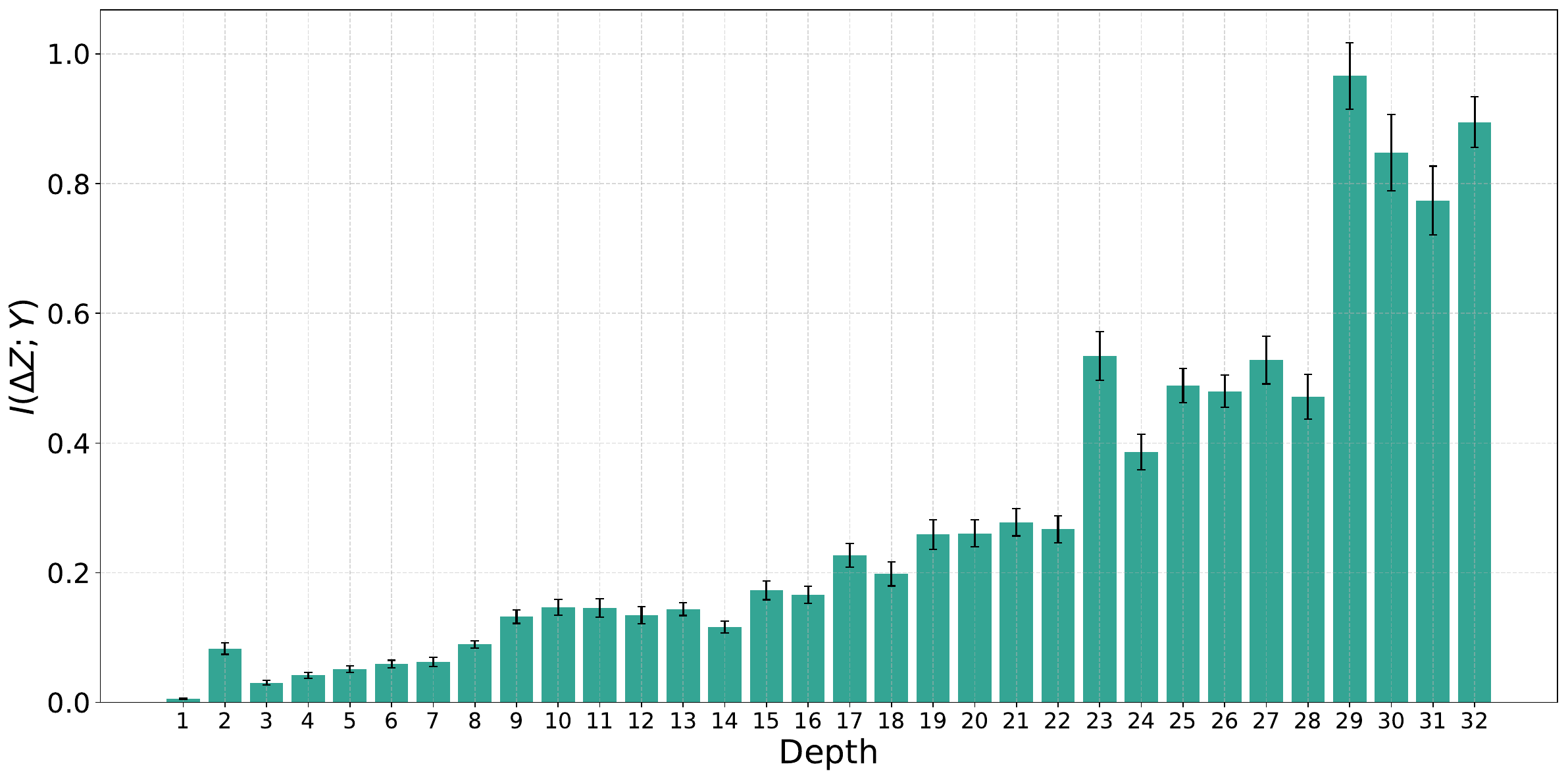}
        \caption{LLaMA-3.1-8B on Synthetic}
    \end{subfigure}
    \caption{Incremental information gain $I(\Delta Z;Y)$ across different models and datasets with \textasciitilde96\% CI error bars.}
    \label{fig:app_idzy_full}
\end{figure}

\begin{figure}[h]
    \centering
    \begin{subfigure}[b]{0.3\linewidth}
        \includegraphics[width=\linewidth]{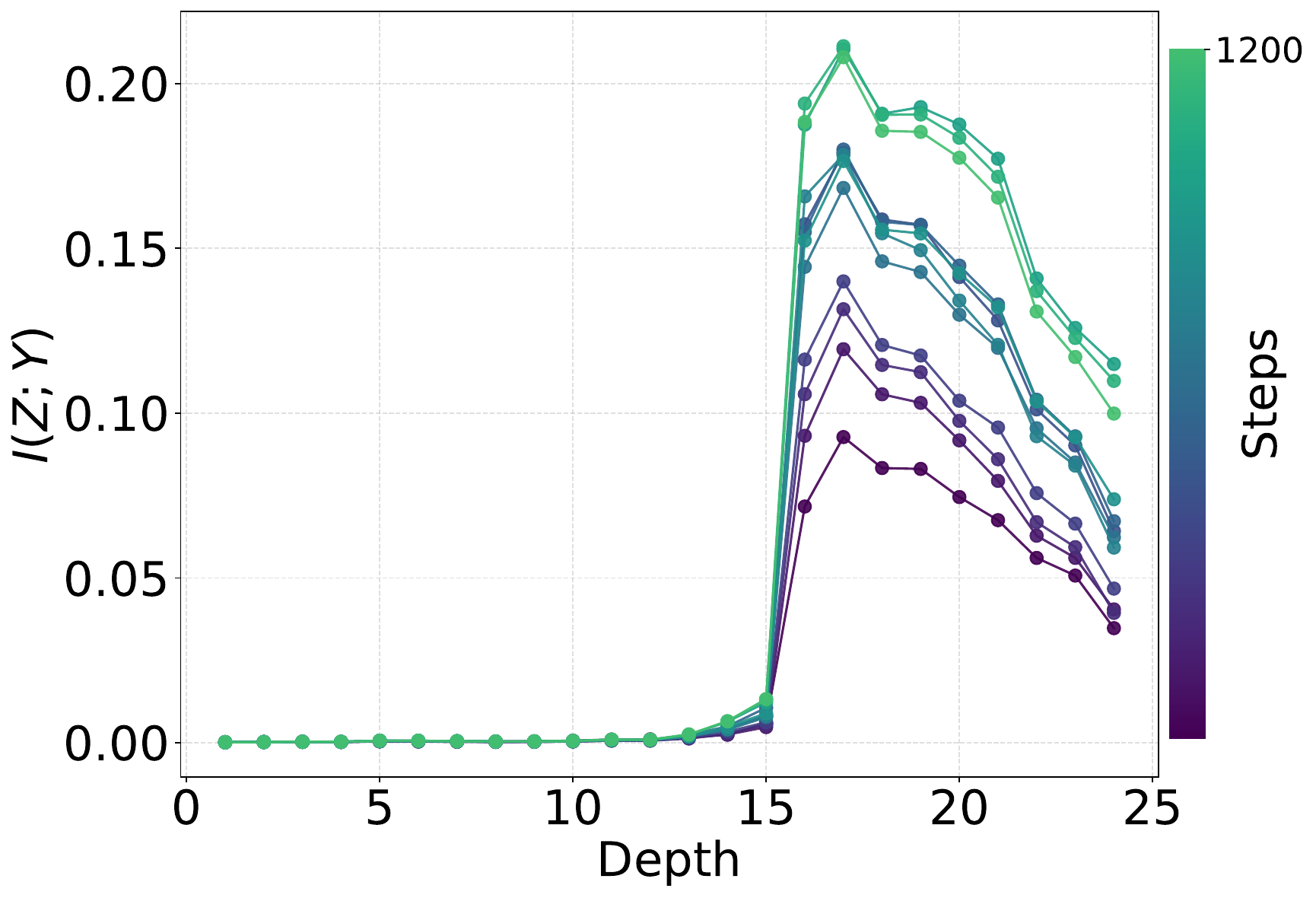}
        \caption{Gaussian Kernel: Qwen-2.5-0.5B on ECQA.}
        \label{fig:sub1}
    \end{subfigure}
    \hfill
    \begin{subfigure}[b]{0.3\linewidth}
        \includegraphics[width=\linewidth]{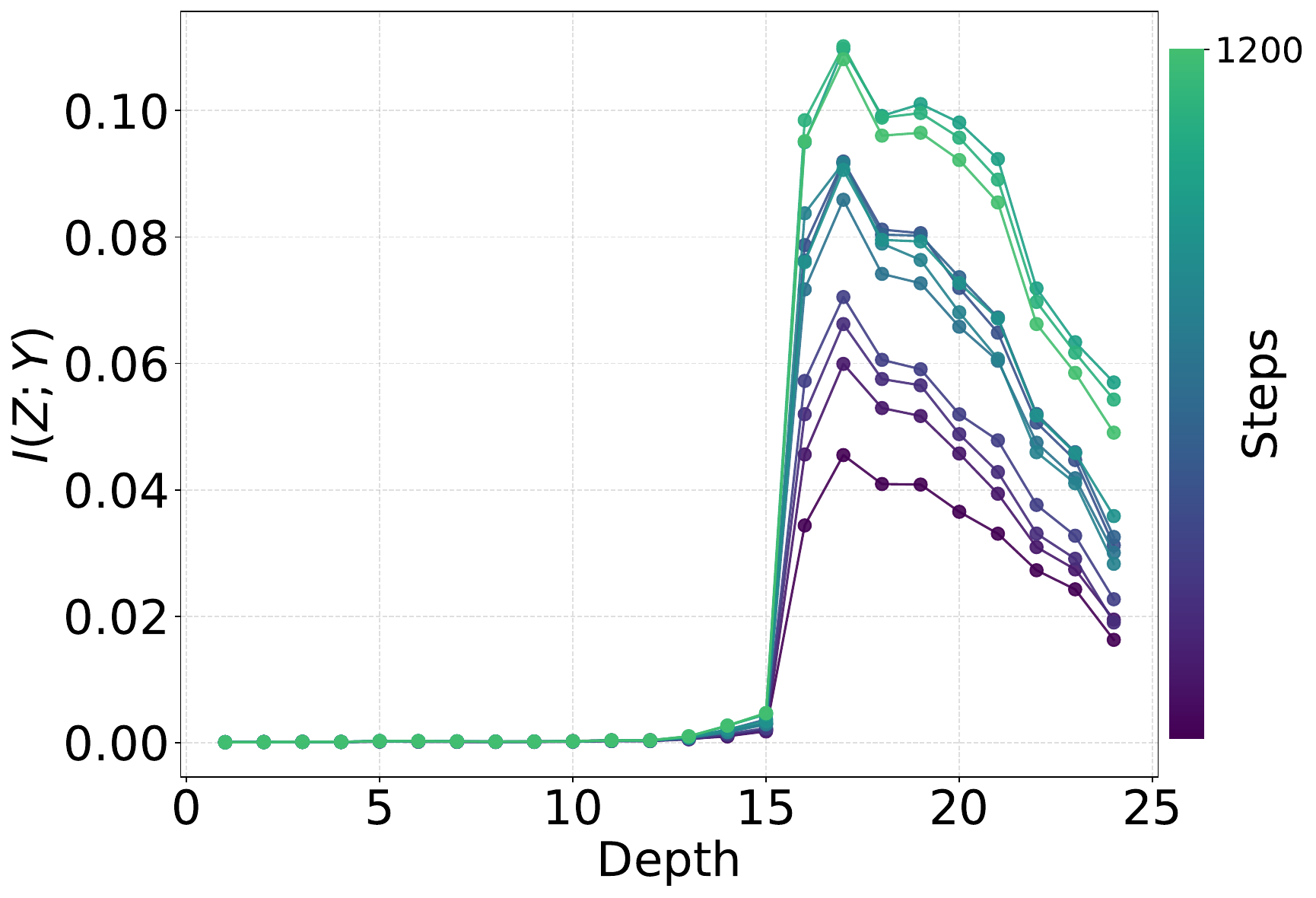}
        \caption{Polynomial Kernel: Qwen-2.5-0.5B on ECQA.}
        \label{fig:sub2}
    \end{subfigure}
    \hfill
    \begin{subfigure}[b]{0.3\linewidth}
        \includegraphics[width=\linewidth]{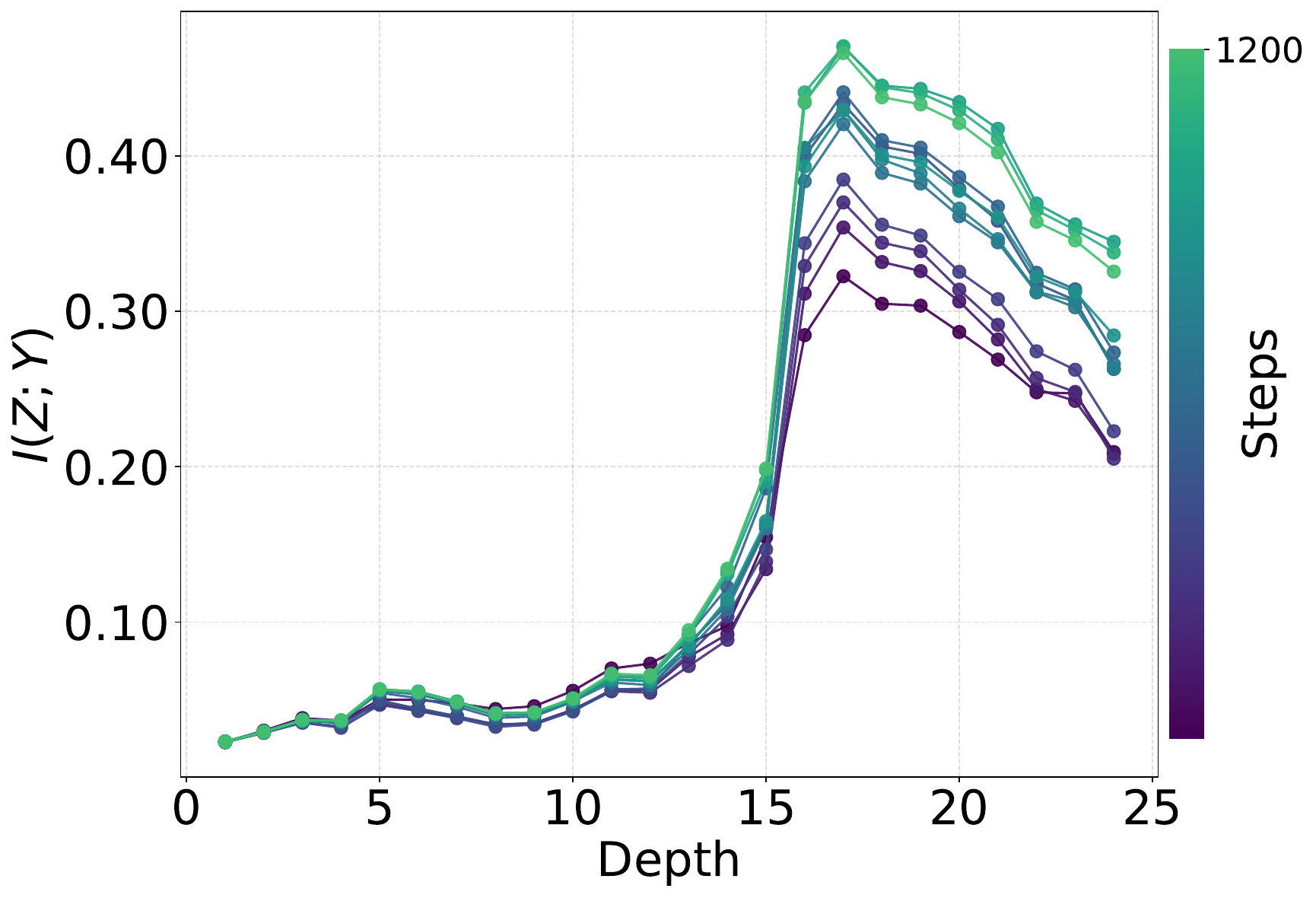}
        \caption{Laplacian Kernel: Qwen-2.5-0.5B on ECQA.}
        \label{fig:sub1}
    \end{subfigure}
    \caption{Different kernel show similar pattern.}
    \label{fig:app_kernel}
\end{figure}

\paragraph{Kernel ablation}
To assess robustness with respect to kernel selection, we additionally test the Laplacian and Polynomial kernels (Figure~\ref{fig:app_kernel}).

\paragraph{Residual scaling} We present the complete set of residual scaling results, detailing the learned $\beta_\ell$ values across all transformer layers.

\begin{figure}[h]
    \centering
    \begin{subfigure}[b]{0.49\textwidth}
        \includegraphics[width=\linewidth]{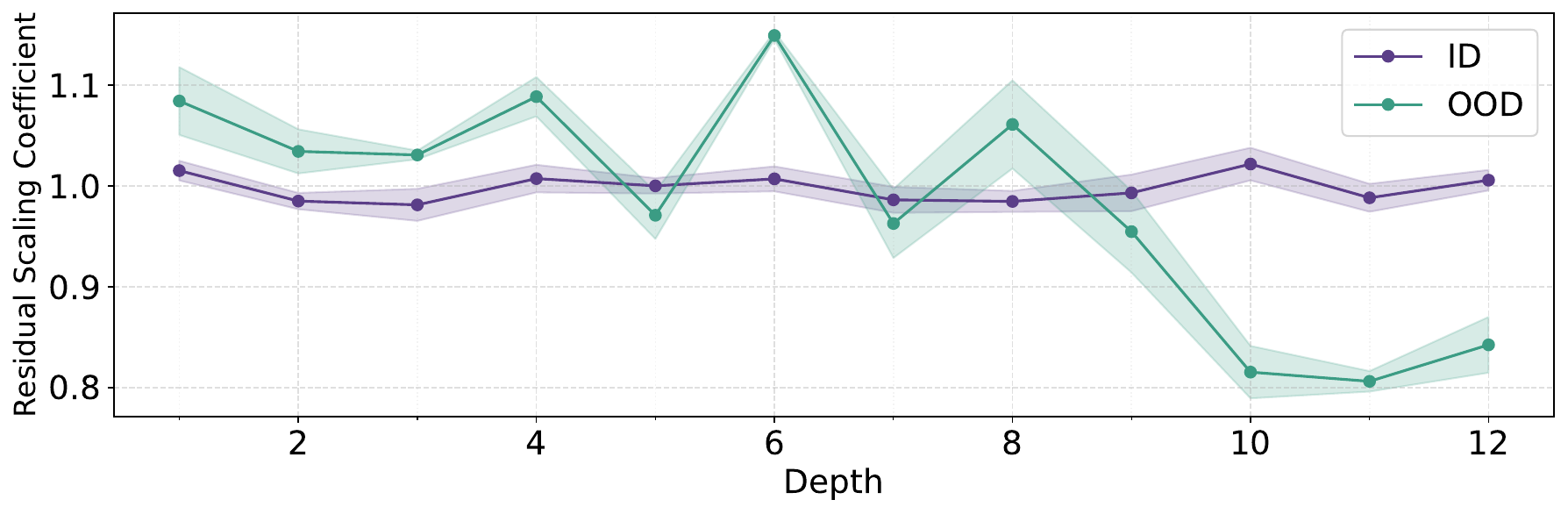}
        \caption{GPT-2 Small on Synthetic Arithmetic}
    \end{subfigure}
    \hfill
    \begin{subfigure}[b]{0.49\textwidth}
        \includegraphics[width=\linewidth]{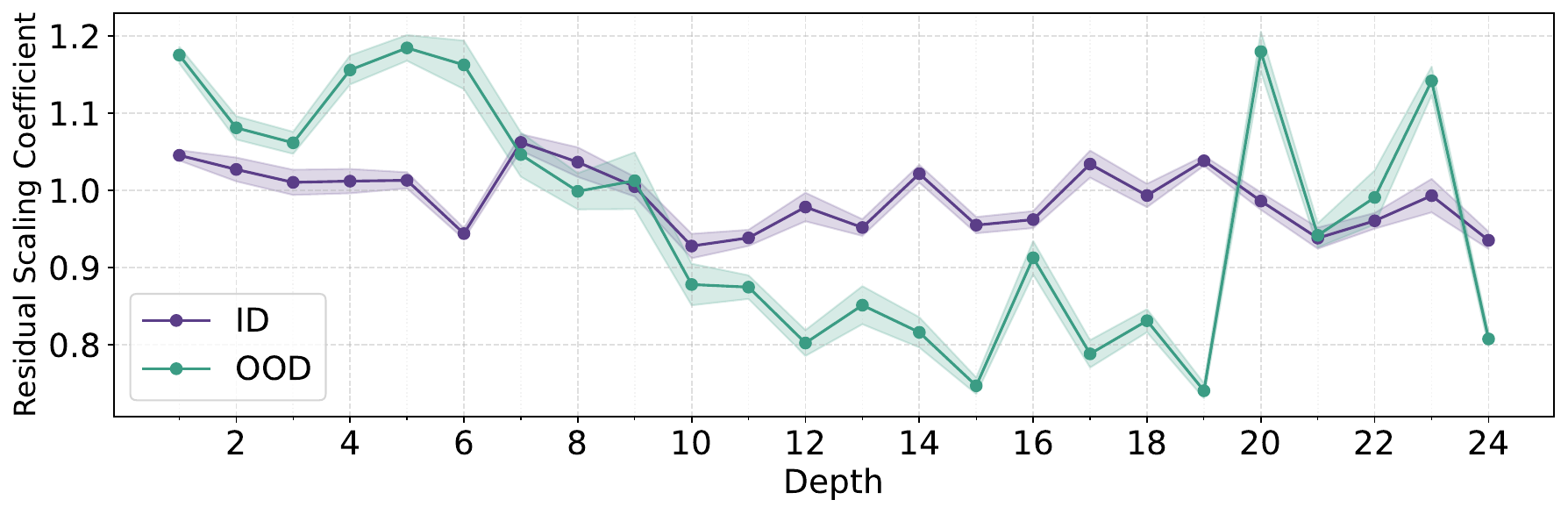}
        \caption{GPT-2 Medium on CLUTRR}
    \end{subfigure}
    \hfill
    \begin{subfigure}[b]{0.49\textwidth}
        \includegraphics[width=\linewidth]{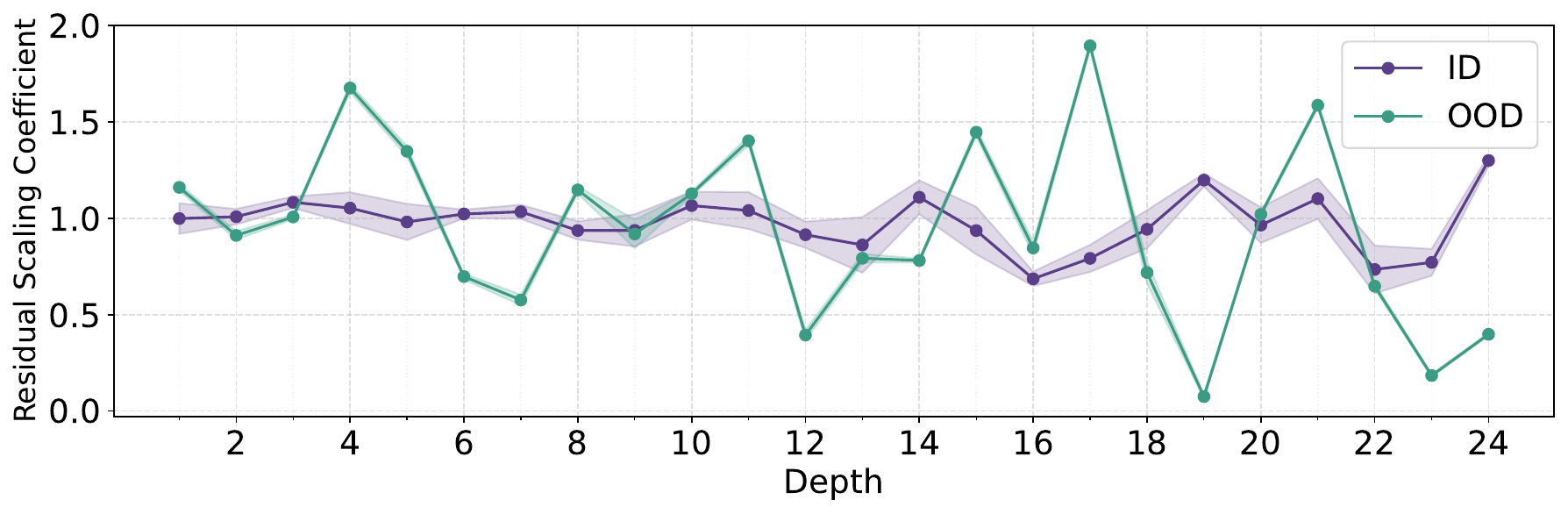}
        \caption{Qwen-2.5-0.5B on ECQA}
    \end{subfigure}
    \hfill
    \begin{subfigure}[b]{0.49\textwidth}
        \includegraphics[width=\linewidth]{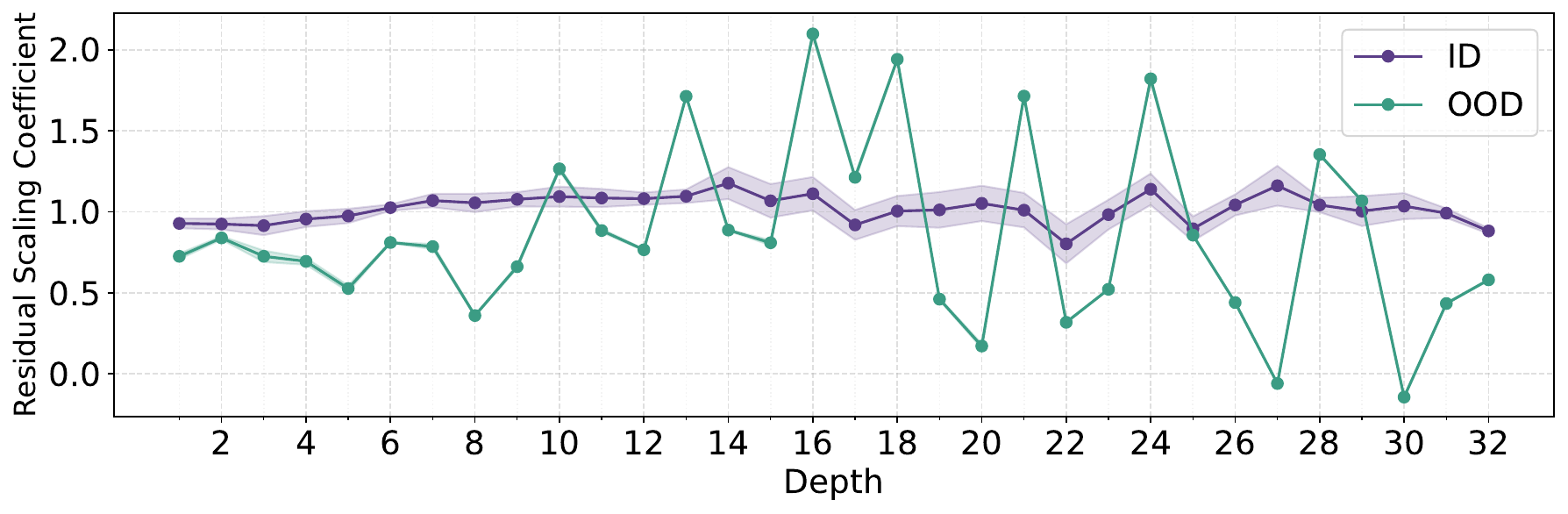}
        \caption{LLaMA-3.1-8B on ECQA}
    \end{subfigure}
    \caption{Residual scaling coefficients $\beta_\ell$ across all transformer layers. Each curve shows the mean across five random seeds, and the shaded region denotes 1-sigma error bar. ID training emphasizes later layers, while OOD training has lower final layer weight.}
    \label{fig:app_beta_full}
\end{figure}

\begin{table}[h]
\begin{center}
\label{tab:train_transfer}
\resizebox{\textwidth}{!}{%
\begin{tabular}{llcc|cc|cc}
\toprule
& & \multicolumn{2}{c}{Before Training} 
& \multicolumn{2}{c}{Train on ID} 
& \multicolumn{2}{c}{Train on OOD} \\
\cmidrule(lr){3-4} \cmidrule(lr){5-6} \cmidrule(lr){7-8}
Model & Dataset 
& ID & OOD 
& ID & OOD 
& ID & OOD \\
\midrule
GPT2        & Synthetic Arithmetic & 100.0$\pm$0.0 & 39.3$\pm$2.1 & 100.0$\pm$0.0 & 39.1$\pm$2.1 & 71.8$\pm$2.6 & 65.4$\pm$3.0 \\
GPT2-Medium & CLUTRR               & 95.8$\pm$0.0  & 28.7$\pm$3.3 & 97.9$\pm$0.0  & 30.4$\pm$4.0 & 97.9$\pm$0.0  & 32.7$\pm$3.4 \\
Qwen-2.5-0.5B & ECQA               & 98.1$\pm$0.3  & 4.2$\pm$0.0  & 98.5$\pm$0.2  & 1.5$\pm$1.8  & 80.0$\pm$2.2  & 14.9$\pm$0.3 \\
LLaMA-3.1-8B & ECQA                & 95.0$\pm$0.4  & 0.0$\pm$0.0  & 95.5$\pm$0.4  & 0.0$\pm$0.0  & 67.6$\pm$2.7  & 21.1$\pm$0.5 \\
\bottomrule
\end{tabular}
}
\end{center}
\caption{Accuracy (mean $\pm$ std) before and after training on ID or OOD data.}
\end{table}

\begin{figure}[h]
    \centering
    \begin{subfigure}[b]{0.48\linewidth}
        \includegraphics[width=\linewidth]{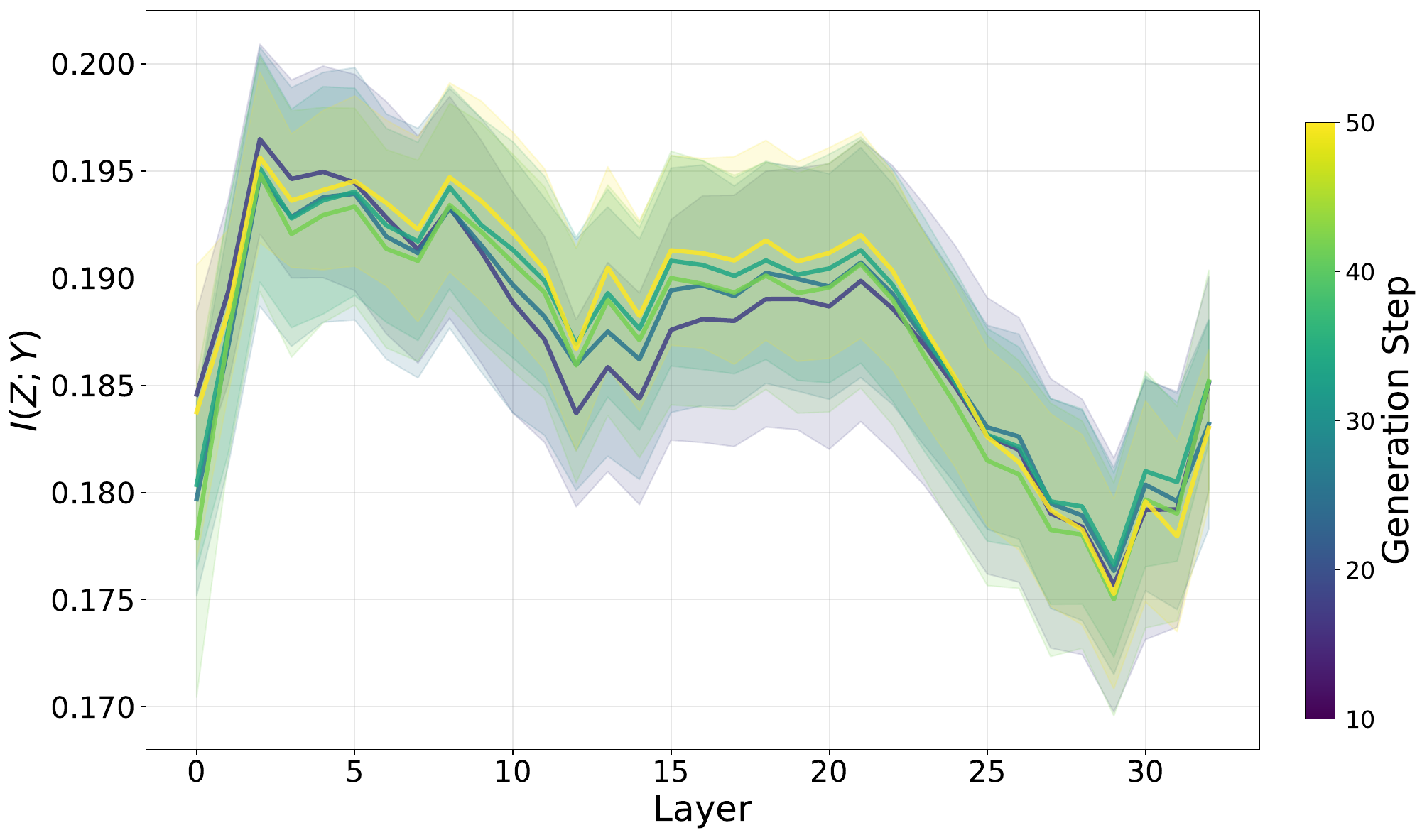}
        \caption{LLaMA on CNN/Daily for 50 steps.}
        \label{fig:sub1}
    \end{subfigure}
    \hfill
    \begin{subfigure}[b]{0.48\linewidth}
        \includegraphics[width=\linewidth]{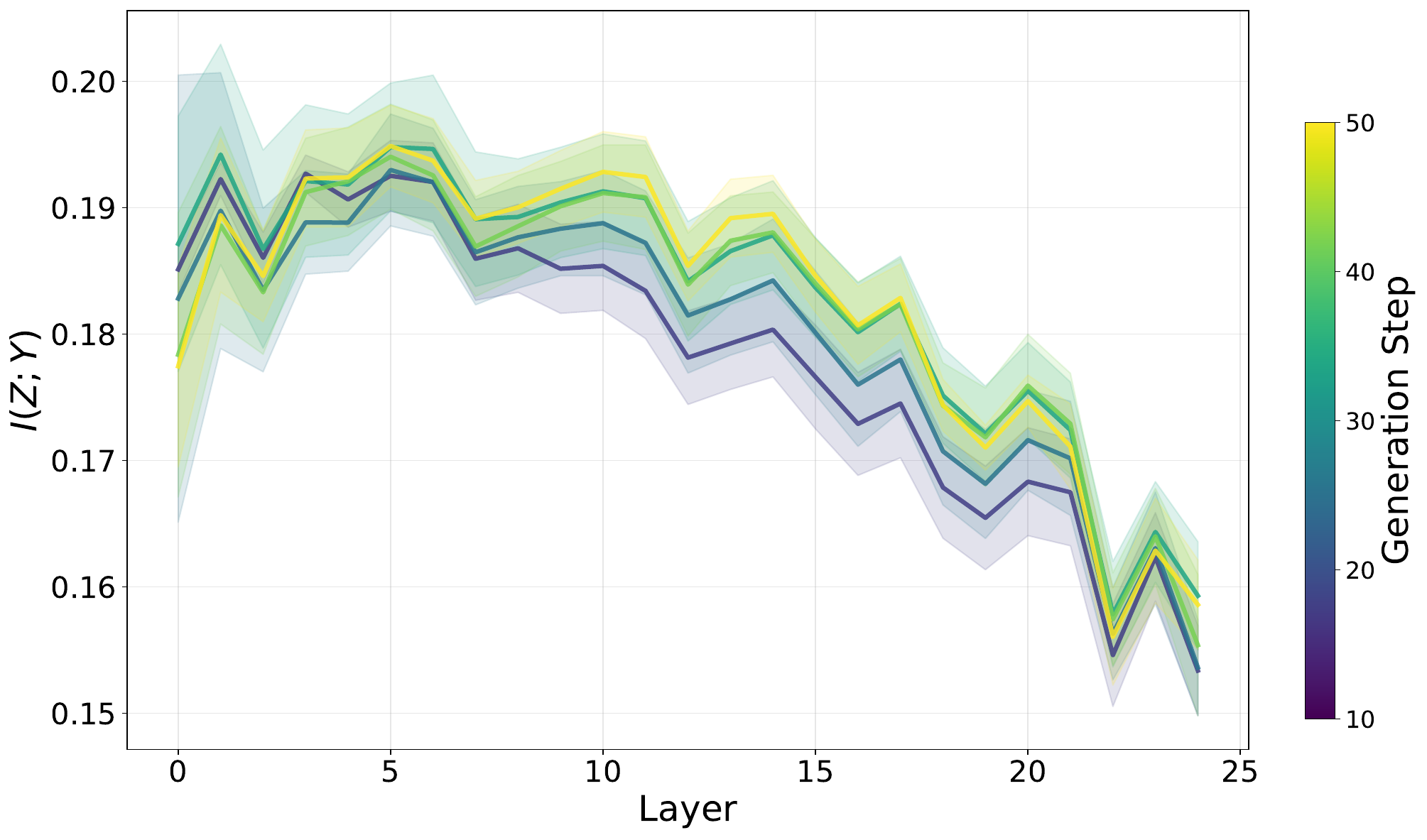}
        \caption{Qwen on CNN/Daily for 50 generation steps.}
        \label{fig:sub2}
    \end{subfigure}   
    \caption{$I(Z_\ell^{(t)} ; Y)$ across layers (x-axis) and generation steps (right color bar), the left y-axis shows mutual information values. Values are plotted every ten steps. Lines denote the mean across three independent runs (seed 42, 123, 1234, temperature 0.7), with shaded regions indicating $\pm 1$ standard deviation.}
    \label{fig:multi}
\end{figure}

\begin{figure}[h]
    \centering
    \begin{subfigure}[b]{0.48\linewidth}
        \includegraphics[width=\linewidth]{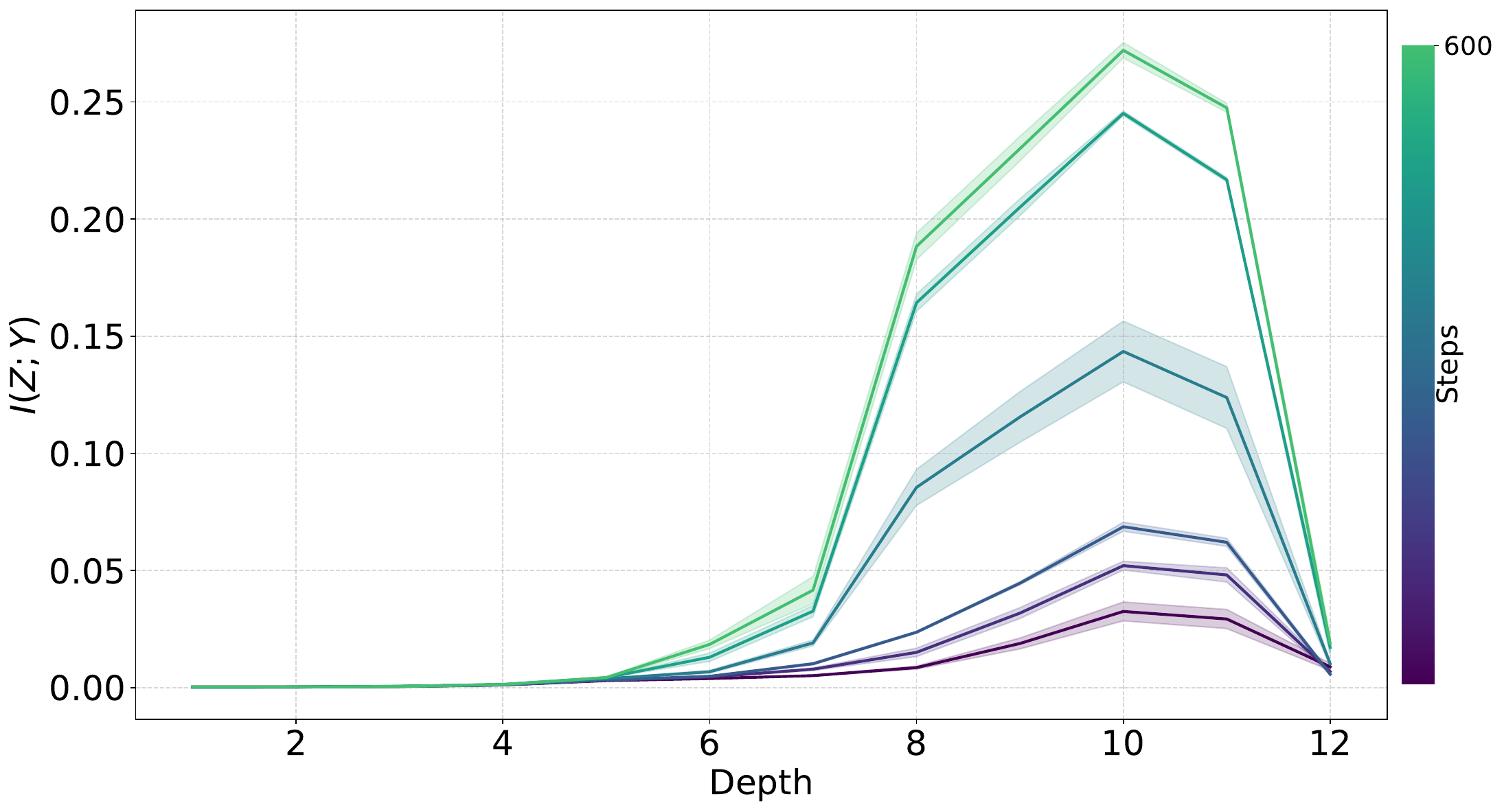}
        \caption{GPT-2 Small on Synthetic}
        \label{fig:sub1}
    \end{subfigure}
    \hfill
    \begin{subfigure}[b]{0.48\linewidth}
        \includegraphics[width=\linewidth]{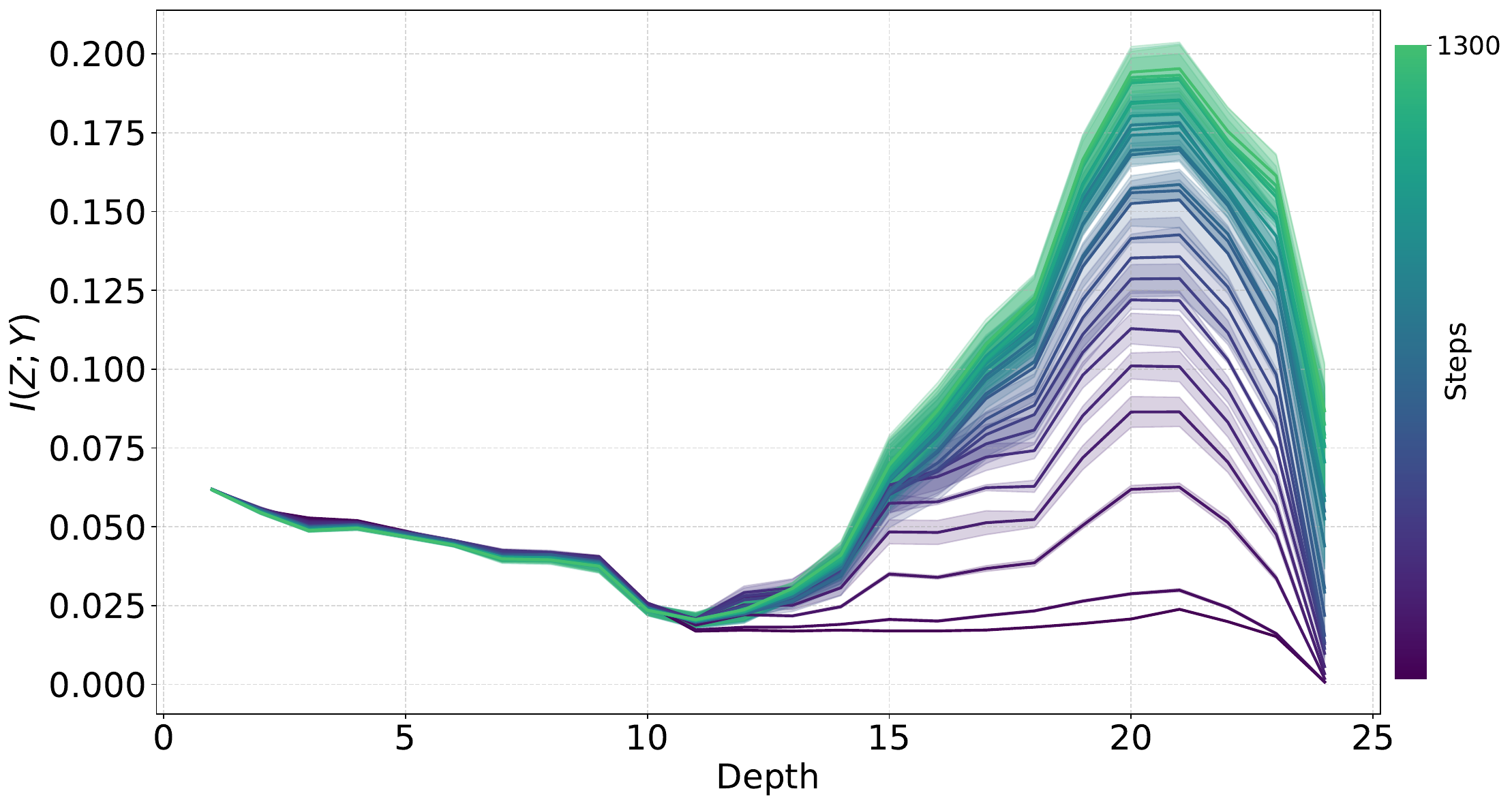}
        \caption{GPT2-Medium on CLUTRR}
        \label{fig:sub2}
    \end{subfigure} 
    \hfill
    \begin{subfigure}[b]{0.48\linewidth}
        \includegraphics[width=\linewidth]{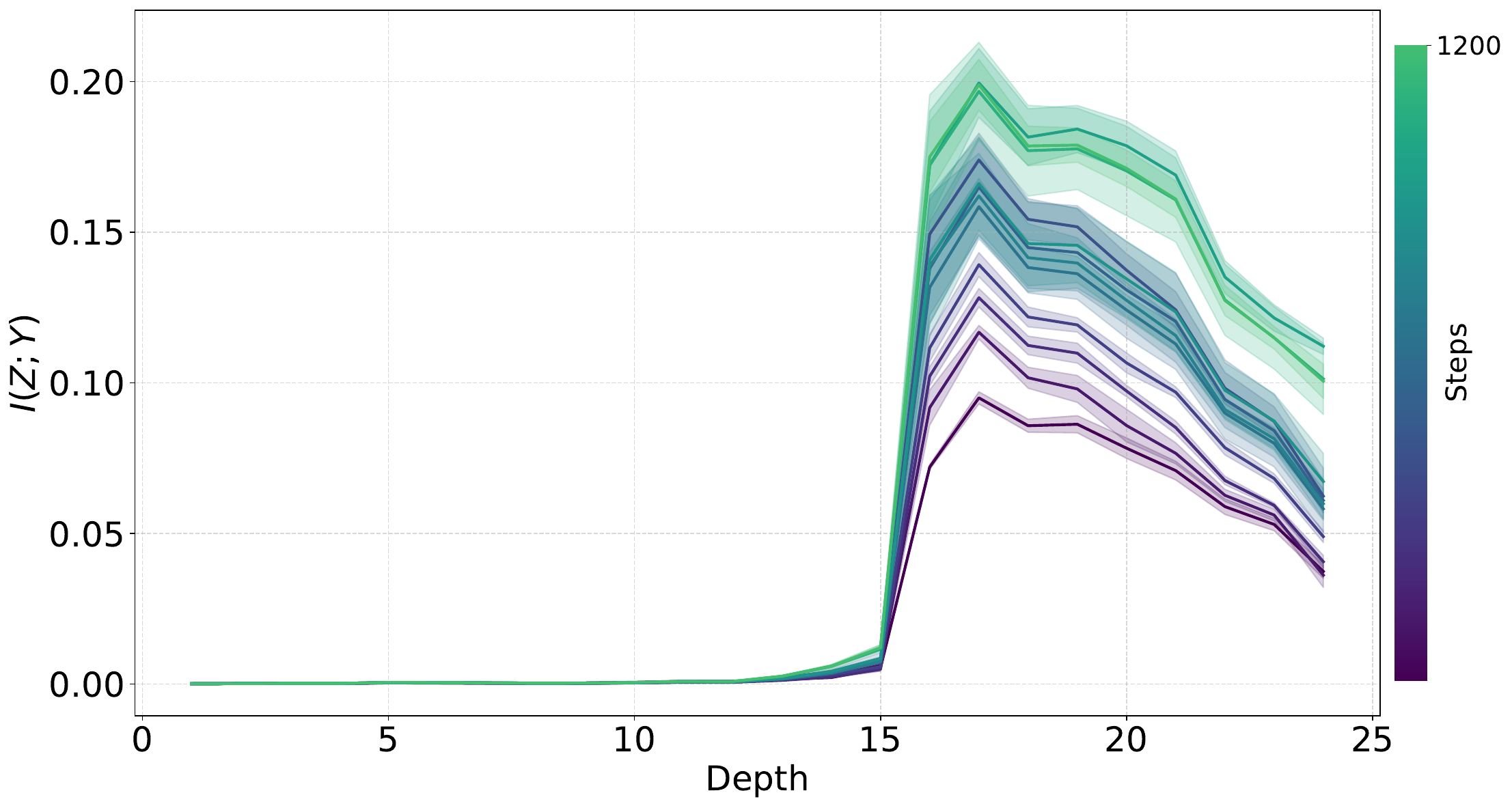}
        \caption{Qwen-2.5-0.5B on ECQA}
        \label{fig:sub1}
    \end{subfigure}
    \hfill
    \begin{subfigure}[b]{0.48\linewidth}
        \includegraphics[width=\linewidth]{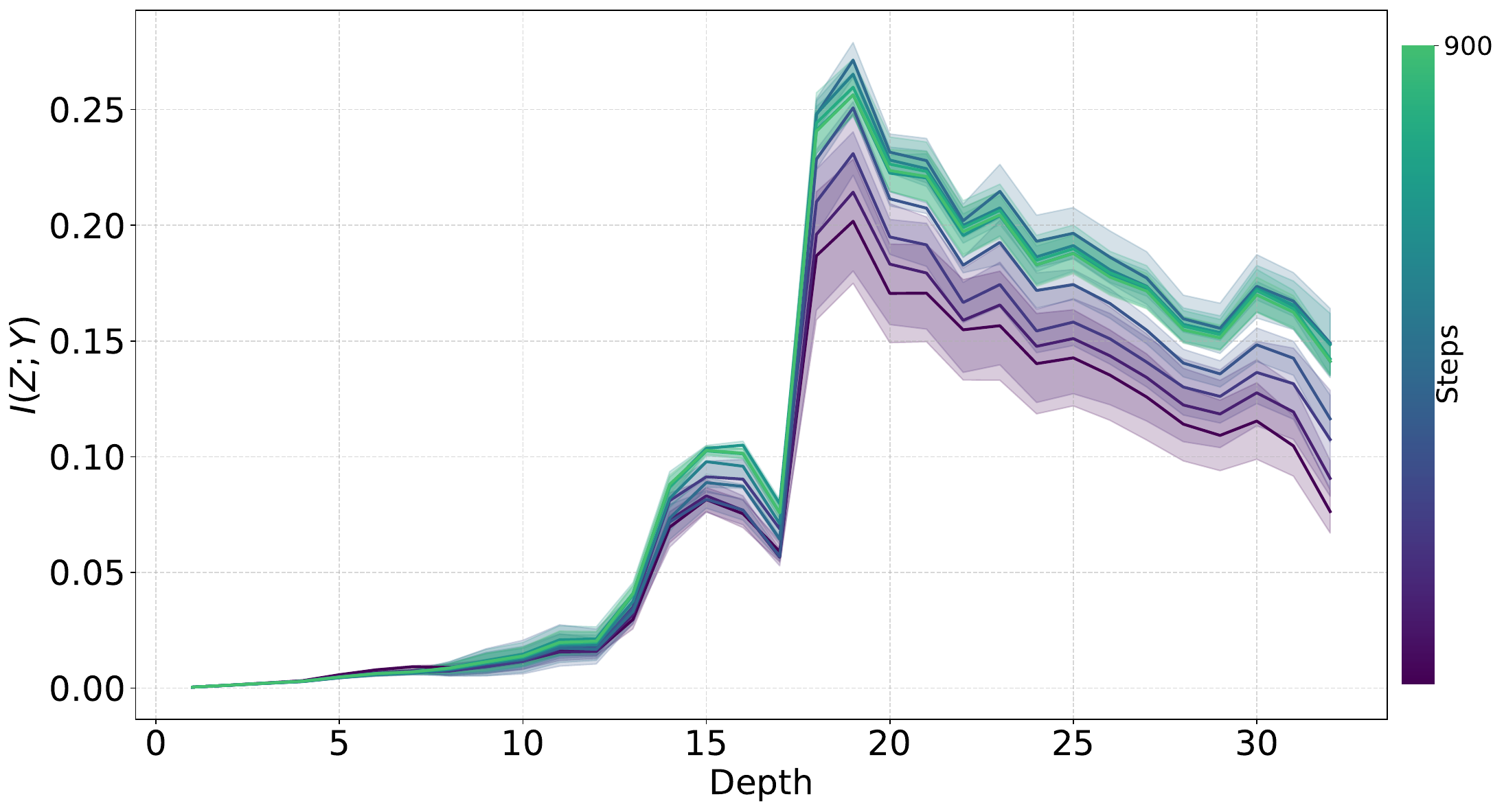}
        \caption{LLaMA-3.1-8B on ECQA}
        \label{fig:sub2}
    \end{subfigure}
    \caption{$I(Z_\ell^{(t)} ; Y)$ across layers (x-axis) with standard deviation across 3 different finetuning seed (42, 123, 1234).}
    \label{fig:multi}
\end{figure}

\end{document}